\documentclass{article}

\usepackage{microtype}
\usepackage{graphicx}
\usepackage{booktabs} % for professional tables

\usepackage{hyperref}

\newcommand\figref{Figure~\ref}

\newcommand\appref{Appendix~\ref}

\usepackage[accepted]{icml2023}

\usepackage{amsmath}
\usepackage{amssymb}
\usepackage{mathtools}
\usepackage{amsthm}
\usepackage{subcaption}

\usepackage[capitalize,noabbrev]{cleveref}

\theoremstyle{plain}

\theoremstyle{definition}

\theoremstyle{remark}

\usepackage{tcolorbox}

\icmltitlerunning{Simulation-based Inference for Cardiovascular Models}

\begin{document}

\twocolumn[
\icmltitle{Simulation-based Inference for Cardiovascular Models}
% Comments Ozan: 
% is it useful to show this "interesting" behavior but then say the simulation does not reflect real data?
% Can we somehow learn why inference on real data does not work from simulation inference, e.g. multi-modality?
% focuse more on the fact that these models are a standard and used in many areas (and the distinction between 3D and 1D)

% Use letters for affiliations, numbers to show equal authorship (if applicable) and to indicate the corresponding author



\icmlsetsymbol{equal}{*}

\begin{icmlauthorlist}
\icmlauthor{Antoine Wehenkel}{equal,comp}
\icmlauthor{Laura Manduchi}{equal,eth}
\icmlauthor{Jens Behrmann}{comp}
\icmlauthor{Luca Pegolotti}{comp}
\icmlauthor{Andrew C.~Miller} {comp}
\icmlauthor{Guillermo Sapiro} {comp}
\icmlauthor{Ozan Sener}{comp}
\icmlauthor{Marco Cuturi} {comp}
\icmlauthor{Jörn-Henrik Jacobsen} {comp}
\end{icmlauthorlist}

\icmlaffiliation{comp}{Apple}
\icmlaffiliation{eth}{ETH Zürich, work done while being at Apple.}

\icmlcorrespondingauthor{Antoine Wehenkel}{awehenkel@apple.com}

\icmlkeywords{Bayesian Inference, Cardiovascular Simulations, Hemodynamics, Machine Learning, Simulation-based Inference}

\vskip 0.3in
]



\printAffiliationsAndNotice{\icmlEqualContribution} % otherwise use the standard text.

\begin{abstract}
Over the past decades, hemodynamics simulators have steadily evolved and have become tools of choice for studying cardiovascular systems in-silico. 
While such tools are routinely used to simulate whole-body hemodynamics from physiological parameters, solving the corresponding inverse problem  of mapping waveforms back to plausible physiological parameters remains both promising and challenging.
Motivated by advances in simulation-based inference (SBI), we cast this inverse problem as statistical inference.
In contrast to alternative approaches, SBI provides \textit{posterior distributions} for the parameters of interest, providing a \textit{multi-dimensional} representation of uncertainty for \textit{individual} measurements. 
We showcase this ability by performing an in-silico uncertainty analysis of five biomarkers of clinical interest comparing several measurement modalities. Beyond the corroboration of known facts, such as the feasibility of estimating heart rate, our study highlights the potential of estimating new biomarkers from standard-of-care measurements. SBI reveals practically relevant findings that cannot be captured by standard sensitivity analyses, such as the existence of sub-populations for which parameter estimation exhibits distinct uncertainty regimes. Finally, we study the gap between in-vivo and in-silico with the MIMIC-III waveform database and critically discuss how cardiovascular simulations can inform real-world data analysis. 
\end{abstract}

\vspace*{-5mm}
\begin{tcolorbox}[colback=white, colframe=black]
This article has an extended version entitled \textit{Leveraging Cardiovascular Simulations for In-Vivo Prediction of Cardiac Biomarkers} \citep[][\href{https://arxiv.org/abs/2412.17542}{Link}]{manduchi2024leveraging}.
\end{tcolorbox}
\vspace*{5mm}

\section{Introduction}
%% CV sims are important and useful
A fine-grained understanding of the human cardiovascular~(CV) system is crucial to mitigating CV diseases. CV models, starting with those of William Harvey~\citep{ribatti2009william}, have seen tremendous progress over the past decades, going from paper calculations~\citep{altschule1938effects, patterson1914regulation} to comprehensive simulators~\citep{updegrove2017simvascular, melis2017gaussian, charlton2019modeling, alastruey2023arterial} that leverage advances in scientific computing. 
Such simulators now provide a personalized description of many aspects of the CV system. They can, for instance, describe cardiac function with 3D models~\citep{baillargeon2014living}, or even simulate hemodynamics in the entire human arterial system~\citep{melis2017gaussian, charlton2019modeling, alastruey2023arterial}. These advances in CV modeling support the development of \textit{personalized} monitoring and treatment of CV diseases, ushering in a new era of precision medicine~\citep{ashley2016towards}.

While whole-body 1D hemodynamics simulators~\citep{melis2017gaussian, charlton2019modeling} establish a clear path from latent physiological variables to measurable biosignals,
% and play an essential role in the study of CV disease~\citep{alastruey2023arterial}.
their use for scientific inquiry, or precision medicine, necessitates solving the corresponding \textit{inverse} problem of inferring latent biomarkers from measurable biosignals. However, this inversion is challenging as the forward model is often specified as a computationally expensive black-box simulator~\citep{manganotti2022modeling}. To add to the complexity, the numerous interactions between parameters lead to convoluted symmetries and ambiguous inverse solutions~\citep{quick2001infinite, nolte2022inverse}. 

Recent works have studied these inverse problems with variance-based sensitivity analysis, highlighting which biomarkers have the most decisive influence on measured biosignals~\cite{melis2017bayesian,schafer2022uncertainty,piccioli2022effect}.
In parallel, machine learning approaches, relying on sophisticated patterns for predicting biomarkers from biosignals, have gained popularity~\citep{chakshu2021towards, jin2021estimating, bikia2021estimation, ipar2021blood,bonnemain2021deep}. While these approaches provide an essential step towards a better understanding of the inverse problem, they do not address the challenges caused by the non-deterministic and multi-modal nature of inverse solutions, as substantiated in Section~\ref{sec:in-silico}.

\begin{figure*}
  \includegraphics[width=\textwidth]{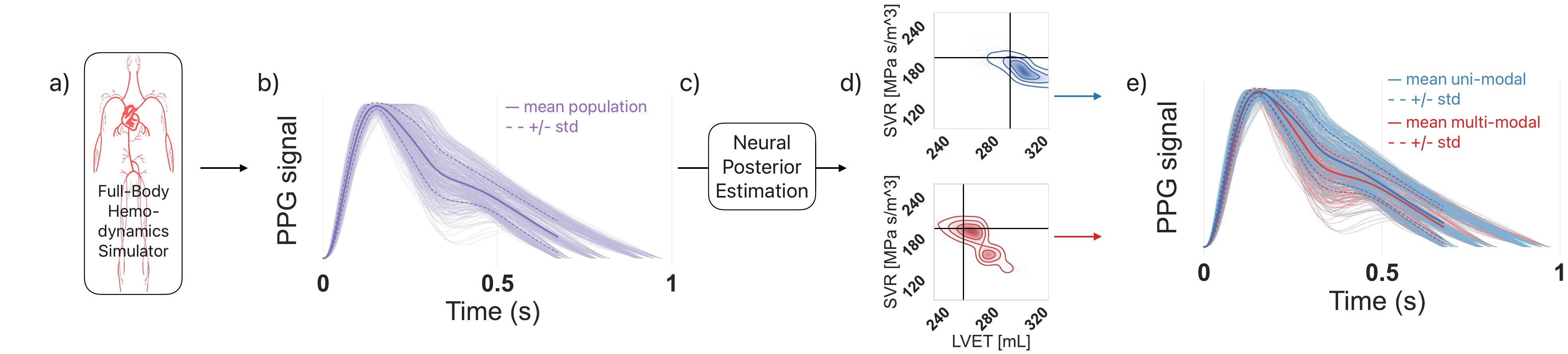}
  \caption{A sketch of SBI for the analysis of 1D cardiovascular models and the corresponding uncertainty analysis performed at an individual level. \textit{Our results indicate that the estimation of left-ventricular ejection time (LVET) and systemic vascular resistance (SVR) from the digital PPG are dependent. There exists a sub-population of measurements for which both quantities are identifiable while there remains a bi-modal uncertainty for the respective other measurements.} \textbf{a}: Simulator of the hemodynamics in the $116$ largest human arteries. 
  \textbf{b}: The measurements generated from simulations following a meaningful prior distribution over the model's parameters.
  \textbf{c}: The neural posterior estimation algorithm learns a surrogate of the posterior distribution of the parameters of interest given the digital PPG.
  \textbf{d}: Two posterior distributions respectively corresponding to an individual measurement from each identified sub-population, highlighting the benefit of uncertainty representation at the individual level.
  \textbf{e}: The sub-sets of measurements corresponding to the two identified sub-population, revealing distinct morphological characteristics in each sub-group.
  }\label{fig:two_populations}
    \vspace{-1.5em}
\end{figure*}

%% SBI extends the information that can be extracted from the inverse problem and answer questions previous methods could not
Motivated by breakthroughs in simulation-based inference~\citep[SBI,][]{cranmer2020frontier, tejero2020sbi}, which has addressed similar challenges in other scientific fields, we go beyond producing point-estimates for such inverse problems and consider instead a distributional perspective supported by neural posterior estimation~\citep{lueckmann2017flexible}. 
As a result, the SBI methodology provides a \textit{consistent, multi-dimensional} and, \textit{individualized} representation of uncertainty and naturally handles ambiguous inverse solutions, as showcased in \figref{fig:two_populations}.

% We start this paper with a condensed background (Section~\ref{sec:SBI-4-CV}) that introduces the reader to the CV model considered and SBI. In Section~\ref{sec:results}, we provide an intuitive overview of the various results we can obtain with SBI.  Section~\ref{sec:methods} provides a more detailed technical review of both the CV model and the SBI implementation choices made. Finally, Section~\ref{sec:discussion} concludes with a thorough discussion on the projected impact of SBI to CV modelling.

\section{Background on Hemodynamics and SBI}\label{sec:SBI-4-CV}
% \section{Background}\label{sec:SBI-4-CV}

% The increasing complexity of cardiovascular models, particularly whole-body 1D cardiovascular models, accurately captures a wide range of configurations and heterogeneous populations. However, this complexity necessitates a reevaluation of traditional sensitivity analyses, as not all parameters are identifiable or relevant biomarkers. Similar challenges have been addressed in fields like evolutionary biology, particle physics and astrophysics, which have employed statistical inference methods to handle nuisance effects. In particular, SBI, an emerging field of ML, provides tools to perform statistical inference of models defined as simulators. We now provide additional context to the reader unfamiliar with the SBI methodology and unveil details on the 1D CV model considered. 
\subsection{Inverting Whole-body 1D Cardiovascular Simulations}\label{sec:background_CV}

We define a simulator as a forward generative process $g: \Theta \rightarrow \mathcal{X}$ that inputs a vector of parameters $\mathbf{\theta} \in \Theta$ and returns a simulation $\mathbf{x} \in \mathcal{X}$. In scientific contexts, simulators encode complex generative processes influenced by many exogenous factors represented by the parameters $\mathbf{\theta}$. As a consequence, scientific simulators often depend on a large number of parameters and are usually stochastic. In practice, we split the parameters $\mathbf{\theta}=(\phi, \psi) \in\Theta = \Phi \times \Psi$ into variables of direct interest $\phi \in \Phi$ and nuisance parameters $\psi \in \Psi$ that are necessary to run the simulations but are not of direct interest for the downstream task. 
% We distinguish these two subsets of variables and split a parameter into $\theta=(\phi, \psi) \in\Theta = \Phi \times \Psi$.
% : parameters $\theta$ live in the cartesian product of parameters of interest $\phi$ and nuisance parameter $\psi$.

We rely in this work on the simulator from \cite{charlton2019modeling}, which describes the hemodynamics in the $116$ largest human arteries, using the principle of mass and momentum conservation in blood circulation. The model's parameters describe the blood out-flowing the left ventricle, uni-dimensional physical properties of each artery, and a lumped-element model of the vascular beds. Each individual simulation involves solving the corresponding partial differential equations with appropriate numerical schemes~\citep{Melis2017}. In addition, we reduce the gap between simulated and real-world data with a stochastic measurement model. The resulting simulations, shown in \appref{app:sample_generation}, can faithfully describe the heterogeneity of real-world data, as encountered when considering various individuals, biosignals, or measurement scenarios. Section~\ref{sec:model} further motivates and describes the model.

Using this model, we assess the identifiability of the parameters $\phi \in \Phi$ of physiological interest, from a given measurement $\mathbf{x} \in \mathcal{X}$. In the following, we also use the terms biomarkers for physiological parameters of interest and biosignals for measurements.
The biomarkers considered are the heart rate~\citep[HR,][]{kannel1987heart}, the left ventricular ejection time~\citep[LVET,][]{alhakak2021significance}, the average diameter of the arteries~\citep[Diameter,][]{patel2005cardiovascular}, the pulse wave velocity~\citep[PWV,][]{sutton2005elevated}, and the systemic vascular resistance~\citep[SVR,][]{cotter2003role}. These biomarkers are chosen because of their relevance to assess CV health as supported by the provided references~\citep{kannel1987heart, alhakak2021significance, patel2005cardiovascular, sutton2005elevated, cotter2003role}. We consider biosignals that are commonly collected in intensive care units (ICUs) or in medical studies: the arterial pressure waveform (APW) at the radial artery and the photoplethysmograms (PPGs) at the digital and the radial arteries. 

% The effect of nuisance parameters $\psi$ is typically marginalised out~\citep{taper2011evidence}. The measurements considered in this study are signals commonly collected in intensive care units (ICUs): the arterial pressure waveform (APW) at the radial artery and the photoplethysmograms (PPGs) at the digital and the radial arteries. 

We consider a population aged $25$ to $75$ with several free parameters $\theta$ that model heterogeneous cardiac and arterial properties. 
% The population heterogeneity together with the fact that all parameters are considered unknown at inference time, lead to challenging inverse problems. 
In this context, a consistent representation of the solution's uncertainty is key, in order to capture \textbf{1.} the effect of nuisance parameters, responsible for the forward model stochasticity; \textbf{2.} the symmetries of the forward model, leading to non-unique inverse solutions; \textbf{3.} the lack of sensitivity, magnifying small output uncertainty into high input uncertainty; and \textbf{4.} the heterogeneity of the population considered, leading to distinct uncertainty profiles.

\subsection{Simulation-based Inference~(SBI)} 
% Over the last years, SBI has become an active subject area in machine learning~\citep{lueckmann2021benchmarking, hermans2021averting, glockler2022variational} with applications in astrophysics~\citep{delaunoy2020lightning, dax2021real, wagner2023images}, particle physics~\citep{brehmer2021simulation}, robotics~\citep{marlier2023simulation}, and many others~\citep{luckmann2022simulation, hashemi2022simulation, tolley2023methods, avecilla2022neural}. 
% With the increasing complexity of CV models, there arises a pressing need for a methodology that can effectively handle nuisance effects and the non-uniqueness of solutions to the inverse problems considered. 
SBI~\citep{cranmer2020frontier} has established itself as an essential tool in various domains of science that rely on complex simulations, e.g., in astrophysics~\citep{delaunoy2020lightning, dax2021real, wagner2023images}, particle physics~\citep{brehmer2021simulation}, neuroscience~\citep{luckmann2022simulation, linhart2022neural},  robotics~\citep{marlier2023simulation}, and many others~\citep{hashemi2022simulation, tolley2023methods, avecilla2022neural}. 
SBI extends statistical inference to statistical models defined implicitly from a simulator, such as the one defined in the previous section. By design, SBI methods aim to provide a consistent representation of uncertainty as demanded by the four requirements listed in the previous paragraph.

While possible, applying statistical inference to simulators is challenged by the absence of a tractable likelihood function. Abstracting all sources of randomness within the nuisance parameters $\psi$, the likelihood is implicitly defined as
$p(\mathbf{x} \mid \phi) \propto \int \mathbb{1}_{\mathbf{x}}\bigl( g\left(\psi, \phi \right) \bigr) p(\psi) d \psi, $ where $\mathbb{1}_{\mathbf{x}}$ is the indicator function for the singleton $\{\mathbf{x}\}$. Considering that only forward simulation is possible, as is the case for the simulator considered in this work, the integral is intractable.

% Given that most simulators only allow sampling from the corresponding distribution $p(\mathbf{x} \mid \phi)$, but not evaluating it, as the latter operation requires nuisance parameters marginalisation, which is intractable in most cases. 

% The success of SBI lies in its ability to to port the rigour of statistical analyses to stochastic simulators by leveraging modern ML methods. 

% The apparent intractability of the likelihood function makes statistical inference over such models challenging. Nevertheless, 
SBI algorithms leverage modern machine learning methods to tackle inference in this likelihood-free setting~\citep{lueckmann2021benchmarking, hermans2021averting, glockler2022variational}. SBI algorithms are broadly categorized as \textit{Bayesian} vs \textit{frequentist} and \textit{amortized} vs \textit{online} methods. In contrast to frequentist methods that only assume a domain for the parameter values, Bayesian methods rely on a prior distribution that encodes a-priori belief and target the posterior distribution, modeling the updated belief on the true parameter value after making an observation. As the prior distribution gets uninformative or the number of observations grows, both methods eventually yield the same inference results, the remaining difference being the interpretation of probabilities as a representation of belief or of intrinsic stochasticity. While online methods focus on a particular instance of the inference problem, amortized methods directly target a surrogate of the likelihood~\citep{vandegar2021neural}, the likelihood ratio~\citep{cranmer2015approximating, hermans2020likelihood}, or the posterior~\citep{lueckmann2017flexible,papamakarios2016fast} that is valid for all possible observations. 

Amortized methods are particularly appealing when repeated inference is required, as new observations can be processed efficiently after an offline simulation and training phase. In this work, we also perform repeated inference over the population of interest and have access to a properly defined prior distribution. Thus, we rely on neural posterior estimation~\citep[NPE,][]{lueckmann2017flexible}, a \textit{Bayesian} and \textit{amortized} method, which learns a surrogate of the posterior distribution $p(\phi \mid \mathbf{x})$ with conditional density estimation. Section~\ref{sec:SBI} details the NPE algorithm and Bayesian uncertainty analysis.

 % Although our experiments focus on 1D simulations, the methodology presented here should benefit the analysis of all categories of cardiovascular models and will provide similar conclusions when models are consistent. Section~4\ref{sec:model} provides additional details on 1D hemodynamics models and motivate it over 0D or 3D models.

% We are interested in finding the best way to analyse whole-body CV simulations. We demonstrate that, SBI, as a mature field of machine learning with impactful applications in physics and many other fields, 
% This section provides a background on the application of simulation-based inference for mining CV simulations and enabling precision medicine.
% In Section~3\ref{sec:in-silico}, we analyse a virtual population of healthy individuals aged $25$ to $75$ from \cite{charlton2019modeling} with SBI~\cite{cranmer2020frontier}. Then, we study the transfer of in-silico results to in-vivo, in Section~3\ref{sec:in-vivo}. 

\begin{figure*}
    \centering
    \includegraphics[width=1.\textwidth]{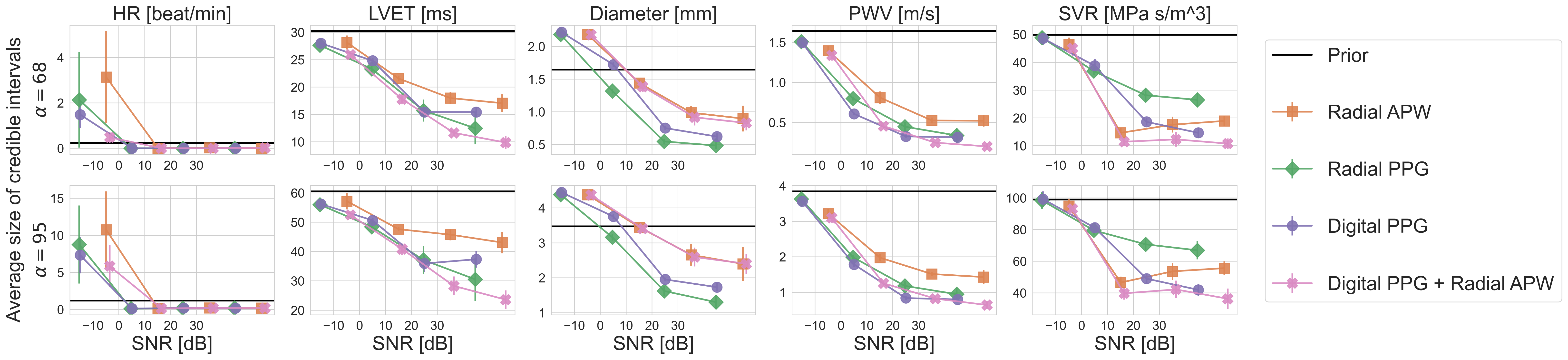}
    \caption{Average size of credible intervals over the test population for credibility levels $68\%$ and $95\%$ of the learned posterior distributions. The x-axis denotes the signal-to-noise-ratio (SNR) for different types of measurements. Results are averaged over five training instances, the vertical bars report one standard deviation. The results are directly expressed in the physical units, in squared brackets, of the parameter considered. \textit{On simulated data, as the SNR increases, the posterior deviates from the prior distribution and the uncertainty over the parameters decreases. The information gain depends on the parameter and measurement considered, no measurement is uniquely better than all others.}}
    \label{fig:identifiability_analysis}
    \vspace{-1.5em}
\end{figure*}

\section{Results}\label{sec:results}
% In this section, we analyse the posterior distributions obtained with SBI. We rely on the neural posterior estimation~\citep[NPE,][]{papamakarios2016fast, lueckmann2017flexible} algorithm to learn a normalizing-flow (NF) based conditional density estimator for each measurement of interest. Compared to other SBI algorithms, NPE is straightforward to train and directly enables evaluating and sampling from the distribution $p(\phi \mid \mathbf{x})$.  
Our experiments consider both in-silico and in-vivo scenarios.  
We split the simulation dataset from \cite{charlton2019modeling} into train ($70\%$), validation ($10\%$), and test sets ($20\%$) at random. All results reported are on the test set for the NPE models that maximize the validation likelihood, error bars report the standard deviation over five training instances. The dataset considered assumed various dependencies between age and most parameters. We prevent age to confound our analysis by explicitly conditioning both posterior and prior distributions on age, and averaging out age from our results. 
% Section~\ref{sec:methods} details further the NPE algorithm and the test metrics.

\subsection{In-silico analysis}\label{sec:in-silico}
% exemplified by successful inference of heart rate without the need of any real training data. However, there remains a non-negligible gap with real data which needs to be closed to allow accurate estimation of the remaining parameters of interest.
\paragraph{SBI enables comprehensive population-level uncertainty analyses.}

Figure~\ref{fig:identifiability_analysis} shows the average size of credible intervals extracted from various posterior distributions of the parameters of interest. With these results, we can compare the identifiability of several parameters given various measurement modalities and levels of noise. We can even inspect the information content of multiple measurement modalities at the same time, as shown in orange, for the Digital PPG and Radial APW. 
% A few examples of the simulated waveforms are provided in \appref{app:sample_generation}. 
To study the robustness of inverse solutions, we consider an additive Gaussian noise model with five noise levels. As a prerequisite for a meaningful analysis, we need to consider inference that is well statistically calibrated, which is the case here as observed in \appref{app:calib_mae}. 
% By construction, for a given measurement modality and noise level, the size of credible intervals~(SCI) with a credibility level equal to $95\%$ are larger than for $68\%$. 
Assuming consistent calibrations, observing the size of intervals as a function of SNR and comparing them to the intervals of the prior distribution quantifies how much information a measurement carries about the biomarkers, as discussed in \appref{app:MI_identifiability}. Section~\ref{sec:metrics} further motivates SCI as an effective metric to study the identifiability of parameters from posterior distributions.

% Overall, there is no unique best measurement. 
Unsurprisingly, the HR is easily identified from all measurements, except for very high levels of noise.
% As the observations are 8-second waveforms it is no surprise that. 
Overall, uncertainty about all parameters reduces significantly as the noise level decreases. This observation indicates that the measurements carry information about all parameters considered which is consistent with the findings of other studies~\citep{melis2017bayesian, charlton2019modeling, charlton2022assessing}. The results also highlight that each measurement has its unique information content. For instance, the digital PPG reveals more about SVR and PWV than the radial PPG. However, it is the opposite for the Diameter for which the Radial PPG is the most informative measurement. 

These results highlight that, similarly to standard sensitivity analyses, SBI enables interpretable assessment of the predictability of biomarkers from biosignals, in-silico, while having additional properties exemplified in subsequent experiments.

\begin{figure}
  \centering
\includegraphics[width=.4\textwidth]{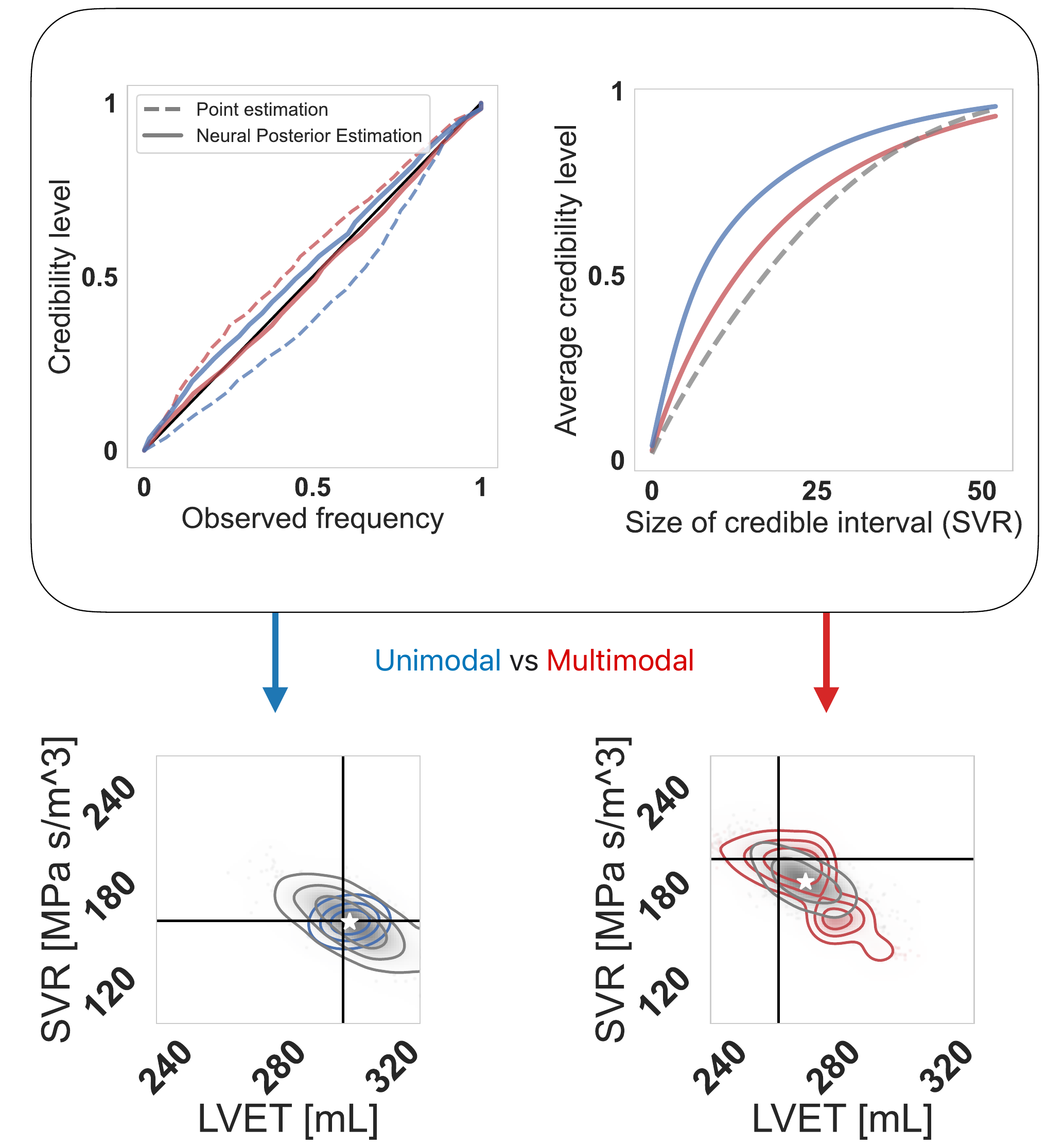}
\caption{Uncertainty obtained from NPE vs. the Laplace's approximation around point estimates, representative of the sensitivity analyses performed in the literature. The colors denote the different populations considered (cf. \figref{fig:two_populations}), the black lines denote the true value of the parameter, and the white star is the point estimate. The left top plot shows that NPE is almost perfectly calibrated whereas the Laplace's approximation is either over-confident or under-confident. The right top plot shows that NPE gives tighter credible intervals compared to Laplace's approximation which yields constant shape. The bottom plot compares the uncertainty representations for two different measurements respectively identified as uni-modal and bi-modal. \textit{Overall this figure shows that NPE (in red and blue) provides a more useful and accurate representation of uncertainty than Laplace's approximation (in gray).}}\label{fig:npe_vs_laplace}
  \vspace{-1.5em}
\end{figure}

\paragraph{SBI enables per-individual uncertainty quantification.}
\figref{fig:npe_vs_laplace} compares the estimation of uncertainty provided by NPE and Laplace's approximation~\citep{MacKay2003Information} around the expectation of the posterior distribution, which is representative of the underlying assumptions made in variance-based sensitivity analyses~(VBSAs). Similarly to VBSAs, Laplace's approximation models uncertainty through a second-order statistic over the population considered.
% \figref{fig:npe_vs_laplace} compares the estimation of uncertainty provided by the posterior distribution obtained with NPE and the uncertainty extracted by Laplace's approximation around the expectation of the posterior distribution. 
Compared to Laplace's approximation, NPE yields tighter and better calibrated credibility intervals. Laplace's intervals tend to be overconfident for measurements that lead to multi-modal posterior distributions, and they are underconfident when the posterior is uni-modal. 
% In comparison, NPE is better calibrated. 
% for all cases although not perfectly, which is expected as NPE is an approximation of the true posterior. 
Furthermore, a point estimator, even with Laplace's uncertainty estimation, will likely assign high density to low-density areas for certain observations, especially if the true posterior is multi-modal. Such inconsistent quantification of uncertainty may mislead downstream decisions. \appref{app:add_exp} showcases the $5$D posterior distributions corresponding to two test examples, which exhibit distinct uncertainty profiles and support further the necessity to use an expressive and observation-dependent quantification of uncertainty.

\figref{fig:two_populations} sketches the use of SBI to study the relationship between the digital PPG and the SVR and LVET. The figure highlights distinctive aspects of posterior distributions within the population studied for which we tested multi-modality~\citep{hartigan1985dip}. While the uncertainty about the value of SVR and LVET can be reduced substantially for approximately half of the test population, for the other half, the posterior is multi-modal. In addition, there remains a strong dependence between the two parameters for this second half. In practice, while a point estimator would be reasonable for the first sub-population, it would be a poor guess for the multi-modal sub-population. Multi-modality indicates that only specific sub-regions of the parameter space are credible, which may suffice to inform certain downstream tasks, e.g., detecting high risk zones, which does not necessarily require knowing the parameter's value exactly. The possibility to perform such fine-grained analysis is only possible with an accurate quantification of uncertainty at the individual level, as offered by SBI.

These results demonstrate that a consistent, multi-dimensional, and individualized representation of uncertainty, as obtained with NPE, yields essential insights from the hemodynamics simulator that are left unnoticed by VBSAs. Multi-modality and dependencies, as observed in \figref{fig:two_populations}, highlight the presence of symmetries in the forward model and individualized uncertainty profiles enable us to stratify the population based on this criterion.

\subsection{In-vivo analysis}\label{sec:in-vivo}
Models are never a perfect representation of real-world data~\citep{box1976science}. Misspecification, as it becomes more significant, hampers the practical relevance of insights extracted from a model~\citep {white1982maximum}. Thus, it is necessary to understand model misspecification to reason about the real world confidently~\citep{geweke2012prediction, box1976science}. Overcoming misspecification is most effectively achieved by identifying conclusions that are independent of the most critical sources of misspecification rather than aiming for perfect models, which do not exist. For instance, in this work, the simulated waveforms strongly depend on the shape of the boundary inflow condition at the aorta, which is approximated with a simplistic and idealized five-parameter descriptions, misspecified for most practical cases. Nevertheless, this description represents accurately the relationship between the length of a beat and the HR. Thus, insights that rely mainly on the beat length of the inflow waveform generalize well to real-world data. However, considering more complex aspects of the model, which reveal more unexpected and intricate relationships between parameters and simulated observations, increases misspecification and identifying definitive conclusions becomes challenging. We further discuss the problem of misspecification in Bayesian inference in Section~\ref{sec:misspecification_Bayesian} 

\begin{figure}
    \centering
    \includegraphics[width=.45\textwidth]{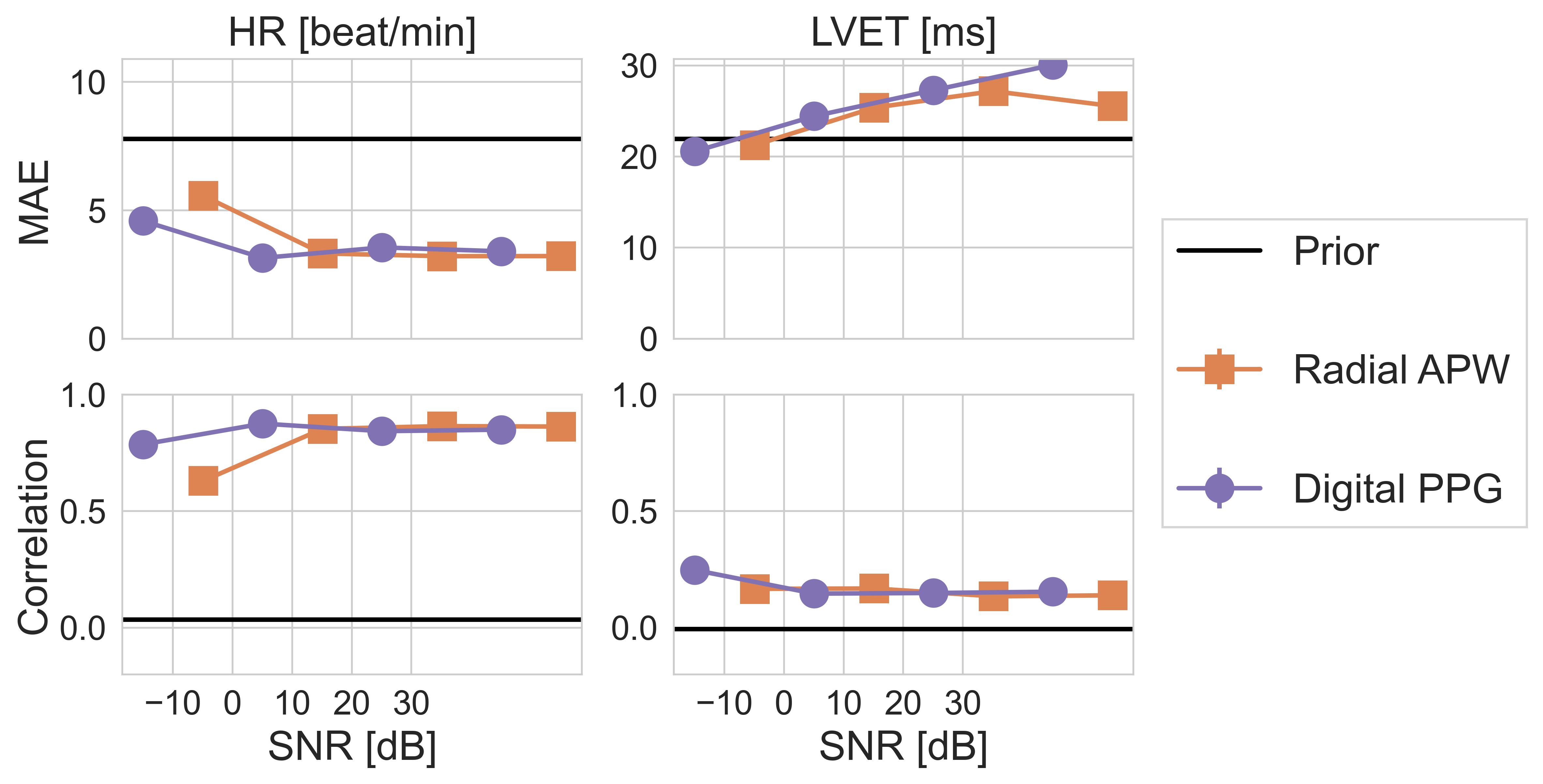}
    \caption{Mean absolute error (MAE) and correlation between the labels and point estimates extracted from the posterior distributions trained for different SNR values. The LVET's performance is compared to the predictions of a prior distribution conditioned on age and HR. \textit{The features predicting HR generalizes better than the one for LVET. The HR MAE decreases with decreasing SNR, indicating the posterior gains robustness to misspecification with decreasing SNR. The features extracted for LVET do not generalize to real-world data but seems to inform more than only age and HR as the posterior's correlation is higher than the prior one. }}
    \label{fig:MIMIC_results}
    \vspace{-1.5em}
\end{figure}

\paragraph{MIMIC-III results.}
In Figure~\ref{fig:MIMIC_results}, we assess the performance of surrogate posterior distributions learned from 1D simulations in predicting HR and LVET using 8-second waveforms from the MIMIC-III dataset~\citep{johnson2016mimic}. Examples of such waveforms are showcased in \appref{app:sample_generation}. As the posterior distributions are uni-modal for the LVET and HR (see, e.g., \appref{app:add_exp}), we focus on point estimates obtained by taking the expectation of the posterior distributions. While we can accurately determine HR by counting the number of beats, assessing LVET is more challenging as there is no gold standard method for obtaining it from PPG or APW. To address this, we use electrocardiograms from MIMIC-III and standard digital signal processing techniques to estimate LVET~\citep{dehkordi2019comparison, alhakak2021significance}. Although this estimation method is not perfect, it serves as a baseline for comparison. We evaluate the mean absolute error (MAE) and correlation between the point estimates and the labels. We must carefully take into account that HR and LVET are negatively correlated in healthy populations, and hence in the prior distribution considered in-silico. We prevent this potentially spurious correlation from corrupting our analysis by explicitly conditioning both prior and posterior distributions on HR and then averaging out the effect of HR.

We reduce prior misspecification (see discussion in Section~\ref{sec:misspecification_Bayesian}) by restricting our analysis to segments that fall within the support of the prior distribution, specifically $\text{HR} \in [ 60, 90]$ and  $\text{LVET} \in [ 230, 330]$. This filtering leaves 547 patients and one measurement per patient. In Figure~\ref{fig:MIMIC_results}, we observe successful transfer of posterior distributions to real-world data for HR but not for LVET. The MAE of LVET approaches that of the prior distribution, indicating limited improvement. However, the remaining correlation between the predicted and real LVET values suggest a partial transfer of information.

The uncertainty analysis in Figure~\ref{fig:identifiability_analysis} aligns with the in-vivo results, showing that HR estimation performs steadily well if SNR is higher than 5dB. At the same time, in-silico and in-vivo results are contradictory for the LVET. In-vivo, the best transfer occur at high noise levels, suggesting that the LVET effect is significantly misrepresented. Investigating and alleviating this misspecification with the appropriate modifications to the model might be crucial to successfully transferring findings from in-silico to in-vivo. This iterative process of \textbf{1.} model analysis, \textbf{2.} real-world experimentation, \textbf{3.} comparison with observations, and \textbf{4.} model refinement; exemplifies the scientific method. In-silico and in-vivo experiments demonstrate that SBI facilitates more scrutiny in applying the scientific loop to cardiovascular models relying on numerical simulations, in extracting scientific hypotheses from the model (step \textbf{1.}); and comparing theoretical predictions and real-world data (step \textbf{3.}).
% Our results demonstrate the value of SBI for mining scientific hypotheses from in-silico cardiovascular models. 
Furthermore, the posterior distributions obtained with SBI provide a multi-dimensional representation of uncertainty and enable to study insights both at the population and the individual level. These analyses provide insights that go beyond what uni-dimensional and population-aggregated uncertainty and identifiability analyses would conclude. 
% Finally, in Section~3\ref{sec:in-vivo}, results on real-world measurements support a critical discussion of how in-silico analyses can provide insights that generalise to the real world. 

% In summary, by carefully examining the posterior distributions learned from simulations onto real-world data, we can identify aspects of the model that effectively transfer in-vivo and those that require further refinement. If necessary, in-silico analyses may guide the resolution of such misspecification by informing when and which parameters are hinted identifiable by the misspecified model. The next step is to gather the corresponding set of real-world labelled data and resolve the misspecification on this set, e.g. by modifying the noise model. A consecutive uncertainty analysis will tell if these parameters are still identifiable under the new and better-specified model.

% By stratifying the population based on expected levels of error using the posterior distribution, we can validate the transferability of the learned posterior distributions. Additionally, an important aspect left for future work is to propose new noise models that produces posterior distributions whose uncertainty is well-calibrated on real-world signals.

\section{Discussion}\label{sec:discussion}

\paragraph{Sensitivity analysis.}
The typical tool for understanding 1D hemodynamics models is variance-based sensitivity analysis~\citep[VBSA,][]{melis2017bayesian, piccioli2022effect,schafer2022uncertainty}.
% Several studies~\citep{melis2017bayesian, piccioli2022effect,schafer2022uncertainty} based on variance-based sensitivity analysis~(VBSA) have made essential contributions to a better understanding of 1D hemodynamics models.
In \cite{melis2017bayesian}, authors perform the analysis on a learnt surrogate of the simulator; \cite{schafer2022uncertainty} relies on polynomial chaos expansion; \cite{piccioli2022effect} studies the effects of various cardiac parameters. 
By contrast, Section~\ref{sec:in-silico} shows that SBI-based uncertainty analysis supports the study of a richer set of uncertainty properties compared to VBSA. For instance, SBI quantifies uncertainty for individual measurements and not only on the population level (highlighted in \figref{fig:npe_vs_laplace}).
% Section~3\ref{sec:in-silico} has shown that SBI-based uncertainty analyses similarly support studying properties of 1D hemodynamics models. 
% Compared to these works, a distinctive attribute of SBI is the ability to quantify uncertainty for individual measurements and not only on the population level (highlighted in \figref{fig:npe_vs_laplace}). 
Furthermore, multi-dimensional VBSA poses computational challenges and is thus not studied in the literature, whereas SBI directly addresses this challenge by learning a joint posterior distribution. Finally, the ambiguity of inverse solutions, which SBI highlights with multi-modal posterior distributions, may invalidate conclusions drawn from uni-dimensional VBSAs.

% our analysis In contrast to  uni-dimensional statistics. 
% We have introduced the SBI methodology to analyse CV simulations, enabling multi-dimensional and individual-level uncertainty analyses. 
% Our results in \figref{fig:npe_vs_laplace} emphasised the need to go beyond these classical sensitivity and uncertainty analyses as neglecting the true complexity of the statistical model may lead to conclusions that are inconsistent with the forward model considered.

\paragraph{ML-based inversion of CV simulators.}
% Another line of work~\citep{chakshu2021towards, jin2021estimating, bikia2021estimation, ipar2021blood, bonnemain2021deep} use ML to inverse CV simulations. 
ML methods can reveal complex relationships between biomarkers and biosignals that are hidden to VBSAs, especially in the presence of salient nuisance effects, as demonstrated in \cite{chakshu2021towards, jin2021estimating, bikia2021estimation, ipar2021blood, bonnemain2021deep}. 
Nevertheless, ML-based sensitivity analyses~(MLBSAs) that do not rely on an effective representation of uncertainty have limitations similar to the ones highlighted for VBSAs in Section~\ref{sec:in-silico}. For instance, they cannot reveal the ambiguity of inverse solutions, as pointed out in \cite{chakshu2021towards}. 
In addition to providing a better understanding of biomarker predictability, MLBSA often aims to address data scarcity, a common challenge for ML in CV applications, with simulated data~\citep{chakshu2021towards, jin2021estimating, bikia2021estimation, ipar2021blood, bonnemain2021deep}. However, as illustrated in Section~\ref{sec:in-vivo}, model misspecification hinders the direct transfer of predictors learned in-silico to real data. Nevertheless, SBI may reveal more general relationships from the simulations (such as dependencies between biomarkers or sub-population identification), which can inform data collection and design of ML algorithms well-suited for the intended CV application.

\paragraph{Next generation of CV simulators.}
Many research projects build upon the versatility and computational efficiency of full-body 1D hemodynamics, relating biosignals to a growing number of biomarkers~\citep{mejia2022effects, buoso2021personalising, coccarelli2021framework}.
Extensions of these models regularly introduce additional parameters, hence additional sources of uncertainty, which further necessitate a probabilistic perspective on the associated inverse problems.
Another avenue for extending CV models is the joint analysis of biosignals relying on distinct physical processes, e.g., PPG and electrocardiograms. In this context, while combining point estimators consistently is complicated, posterior estimators of each model can be combined easily into a joint posterior distribution. Indeed, assuming a conditional independence  $\mathcal{X} \perp \mathcal{Y} \mid \Phi$ between the two forward models $p(x\mid \phi)$ and $p(y\mid \phi)$ allows rewriting the posterior distribution given the two measurements as the product of each posterior distribution: $p(\phi \mid x, y) \propto p(\phi \mid x) p(\phi \mid y)$.
% \paragraph{SBI and the scientific loop.}
% The main advantage of SBI is its ability to port the rigour and practicality of statistical analyses to complex CV simulators. 
% As Section~4\ref{sec:in-vivo} mentions, statistical inference plays two roles in scientific enquiry. First, it reveals implicit hypotheses that follow from the model, guiding real-world experiments that aim to confirm or reject these implications. Second, analysing the gathered real-world data requires another statistical inference tied to the model considered. Our discussion also highlighted the importance of coping with model misspecification. Contrary to what one might think, the growing complexity of CV models does not especially result in lighter misspecification but can increase it. In this context, SBI is particularly well-positioned to address these challenges as it may directly extend robust Bayesian and frequentist inference strategies~\citep{berger1994overview, huang2023learning, cherief2020mmd} to simulators. SBI should thus provide long-term support to research in CV health.

\section{Materials \& Methods} \label{sec:methods}
% This section details the hemodynamics numerical simulation and formally the NPE algorithm and the metrics used to evaluate the inference results.

\subsection{1D hemodynamics of the human arterial network} \label{sec:model}
The full-body arterial model introduced in \cite{alastruey2012arterial}, on which \cite{charlton2019modeling} relies, describes the arterial pulse wave propagation into $116$ arterial segments, making up the largest arteries of the thorax, limbs, and head. This model is a good compromise between faithfulness to the real-world system and complexity~\citep{alastruey2023arterial}. It enables forward simulating APWs and PPGs at multiple locations, given a set of physiological parameters describing the geometrical and physical properties of the cardiovascular system. Running a simulation takes a few minutes on any standard CPU~\citep{melis2017bayesian}, allowing \cite{charlton2019modeling} to release a dataset of $4374$ simulated healthy individuals aged $25$ to $75$.

Compared to 3D and 0D models, 1D models offer a better balance between expressivity and efficiency. While 1D simulations may be less accurate than 3D models (e.g., they cannot model atherosclerosis as they do not consider wall shear stress), they trade a modest and well-studied decrease in accuracy against much lighter simulation costs~\cite{xiao2014systematic, alastruey2023arterial}. Furthermore, the tractable parameterization and efficient simulation of 3D whole-body hemodynamics remain two open research questions~\citep{pegolotti2023learning, lasrado2022silico}. 
On the other side of the CV modeling spectrum, 0D simulations~\citep{john2004forward, shi2011review} rely on a lumped-element model to describe the relationship between blood flow at one location (e.g., left ventricle outflow) and blood pressure and flow at other locations. In addition to ignoring significant physical effects such as wave propagation and reflection, 0D models are partially parameterized by non-physiological quantities. Generating a representative population, such as the one considered in \cite{charlton2019modeling}, can thus be challenging with these models.

\paragraph{Model description.}
In \cite{alastruey2012arterial}, the authors consider the compartmentalized arterial model made of the following sub-models: \textbf{1.} the heart function; \textbf{2.} the arterial system; \textbf{3.} the geometry of arterial segments; \textbf{4.} the blood flow; and \textbf{5.} the vascular beds. \textit{The heart function} describes the blood volume along time at the aorta as a five-parameter function. \textit{The arterial system} is described as a graph, the heart is the parent root, and then arteries branch out into the body. Every branch of the network represents an arterial segment. Segments are coupled so that the conservation of mass and momentum hold in the complete system. Additionally, the heart function defines the boundary condition on the parent root of the arterial network. The vascular bed describes the boundary condition on the leaf nodes. \textit{The geometry of arterial segments} assumes the segments are axial-symmetric and tapered tubes. Hence, the geometry of each arterial segment can be described using 1D parameters such as radius and thickness of the arterial wall. \textit{The blood flow} in the 1D segments follows fluid dynamics, which depends on the geometry and visco-elastic properties of the arterial wall. \textit{The vascular beds} are modeled using 0D approximations, i.e., the geometrical description is being lumped into a space-independent parametric transfer function.

% Following this high-level overview of the 1D hemodynamics models, we now provide some mathematical details on the core components. 
The main state parameters of whole-body 1D hemodynamics models are the volumetric flow rate $Q(z,t)$, the blood pressure $P(z, t)$, and the vessel cross-sectional area $A(z,t)$ at axial position $z$ and time $t$, in each artery considered. Based on the conservation of mass and momentum, one can derive the partial differential equations (PDEs)
\begin{align}
    \frac{\partial A}{\partial t} + \frac{\partial Q}{\partial z} &= 0 \\
    \frac{\partial Q}{\partial t} + \frac{\partial }{\partial z} \left(  \alpha \frac{Q^2}{A}\right) + \frac{A}{\rho} \frac{\partial P}{\partial z} &= -2 \frac{\mu}{\rho} (\gamma_\nu +2)\frac{Q}{A},
\end{align}
where $\alpha$ is the Coriolis' coefficient, $\mu$ is the blood dynamic viscosity, and $\gamma_\nu$ is a parameter defining the shape of the radial velocity profile. A third relationship of the arterial wall mechanics relates pressure and cross-section area as
\begin{align}
    P(A) = P_{ext} + \beta \left( \sqrt{A} - \sqrt{A_0}\right)
    + \frac{\Gamma}{\sqrt{A}} \frac{\partial A}{\partial t}, \\
     \text{where } \beta = \frac{4}{3} \frac{\sqrt{\pi} Eh_0}{A_0} \text{ and } \Gamma = \frac{2}{3} \frac{\sqrt{\pi} \varphi h_0}{A_0}
\end{align}
respectively denote the elastic and viscous components of the Voigt-type visco-elastic tube law,
$P_{ext}$ is the reference pressure at which the geometry is described by the cross-sectional area $A_0$ and thickness of the arterial wall $h_0$. The elastic modulus $E$ and wall viscosity $\varphi$ characterize the mechanical properties of the wall. In addition to these PDEs, boundary conditions are formulated by coupling each artery segment with the parents and children in the arterial network. For further details, see \cite{Melis2017, charlton2019modeling, alastruey2012arterial}.

The considered 1D hemodynamics model constitutes a complex simulator with many parameters. As described in Section~\ref{sec:SBI-4-CV}, only a subset of these parameters are of direct interest. Other parameters are considered nuisance effects. In addition, we consider a measurement model that generates biosignals similar to the one in MIMIC-III. \appref{app:hemodynamics_model} provides additional details on the parameters distributions and the measurement model considered.

% Some parameters are constant across the population, e.g., the blood viscosity $\mu$ or the Poisson ratio $\nu$. Others, such as the one describing the geometric properties of each arterial segment can vary across subjects. \appref{app:model_parameters} describes the parameters treated as nuisance parameters. It also 
% Overall we can leverage the concept of parameters of interest $\phi$ and nuisance parameters $\psi$ from theoretical physics (see Section~4\ref{sec:SBI}) to structure the parameter space. In this work, this structure is guided by selecting parameters $\phi$ which are relevant to describe the coarse health of the cardiovascular system. 

% \paragraph{Measurements modelling.} We utilise a dataset of $4374$ simulations from healthy individuals aged $25$ to $75$~\citep{charlton2019modeling}. By deriving APW and simulating PPG waveforms from the blood flow and pressure data, we obtain signals corresponding to a single heartbeat of varying lengths. To standardise the analysis, we first generate segments longer than 10 seconds by concatenating the same beat multiple times. Then, we randomly crop time series into 8-second segments. This ensures that the posterior distributions are defined for 8-second segments and accounts for all possible starting positions within the heartbeat. Finally, we introduce a white Gaussian noise to the waveforms to make our analysis less sensitive to the model misspecification. \appref{app:sample_generation} showcases these steps and the resulting waveforms.

\subsection{Simulation-based inference} \label{sec:sbi}
\label{sec:SBI}
\paragraph{Neural Posterior Estimation (NPE).}
% We take a Bayesian approach and assume that a well-motivated prior distribution $p(\phi)$ over the parameters is given. We rely on NPE~\citep{papamakarios2016fast, lueckmann2017flexible}, a popular SBI algorithm. 
As mentioned in Section~\ref{sec:SBI-4-CV}, NPE~\citep{papamakarios2016fast, lueckmann2017flexible} is a Bayesian and amortized SBI algorithm. It trains a parametric conditional density estimator for the parameters of interest $p_{\omega}(\phi \mid \mathbf{x})$ on a dataset $\mathcal{D} := \{(\phi_i, \mathbf{x}_i)\}_{i=1}^N$ of samples from the joint distribution $p(\phi, \mathbf{x}) =  \int p(\phi, \psi) p(\mathbf{x}\mid \phi, \psi) d\psi$. In this work, we rely on a rich class of neural density estimators called normalizing flows~\citep[NF, ][]{tabak2010density, tabak2013family, rezende2015variational, kobyzev2020normalizing, papamakarios2021normalizing}, from which both density evaluation and sampling is possible. 

% The NPE algorithm optimises the parameters of the neural density estimator with stochastic gradient descent on the Kullback-Leibler divergence between the conditional density estimator and the posterior distribution. 
Given an expressive class of neural density estimators $\{p_\omega(\phi \mid \mathbf{x}): \omega \in \Omega\}$, NPE aims to learn an amortized posterior distribution $p_{\omega^\star}(\phi \mid \mathbf{x})$ that works well for all possible observation $\mathbf{x} \in \mathcal{X}$, by solving
\begin{align}
    % & \omega^\star \in \arg\min_{\omega \in \Omega} \mathbb{KL}\left[ p(\phi \mid \mathbf{x}) \parallel p_\omega(\phi \mid \mathbf{x}) \right] \quad \forall \mathbf{x} \in \mathcal{X}  \label{eq:goal_NPE}\\
     & \omega^\star \in \arg\min_{\omega \in \Omega} \mathbb{E}_{\mathbf{x}} \left[ \mathbb{KL}\left[ p(\phi \mid \mathbf{x}) \parallel p_\omega(\phi \mid \mathbf{x}) \right] \right] \label{eq:expectation}\\ 
     \iff & \omega^\star \in \arg\min_{\omega \in \Omega} \int p(\mathbf{x}) p(\phi \mid \mathbf{x}) \left[ \log \frac{p(\phi \mid \mathbf{x})}{ p_\omega(\phi \mid \mathbf{x})} \right] d\mathbf{x} d\phi\\
    \iff  & \omega^\star \in \arg\max_{\omega \in \Omega} \int p(\mathbf{x}) p(\phi \mid \mathbf{x}) \log p_\omega(\phi \mid \mathbf{x}) d\mathbf{x} d\phi\\
    \iff  & \omega^\star \in \arg\max_{\omega \in \Omega} \mathbb{E}_{(\phi, \mathbf{x})} \left[ \log p_\omega(\phi \mid \mathbf{x}) \right]. \label{eq:objective_NPE}
    \end{align}
% \eqref{eq:goal_NPE} indicates the goal of NPE: learning a surrogate $p_{\omega^\star}$ that equals the posterior distribution for all values $\mathbf{x} \in \mathcal{X}$. Under the assumption that the class of functions considered contains the true posterior $p(\phi\mid \mathbf{x})$, the minimisers $\omega^\star$ of \eqref{eq:goal_NPE} and \eqref{eq:expectation} are the same. 

In practice, NPE approximates the expectation in \eqref{eq:objective_NPE} with an empirical average over the training set $\mathcal{D}$ and relies on stochastic gradient descent to solve the corresponding optimization problem. Assuming $\phi \in \mathbb{R}^k$ and unpacking the evaluation of the NF-based conditional density estimator, the training loss is
\begin{align}
    \ell(\mathcal{D}, \omega) 
    % &= \frac{1}{N}\sum_{i=1}^N \log p_{\omega}(\phi_i \mid \mathbf{x}_i)\\
    &= \frac{1}{N}\sum_{i=1}^N \log p_z\biggl(f_{\omega}\bigl(\phi_i; \mathbf{x}_i\bigr)\biggr) + \log \lvert J_{f_\omega}(\phi_i; \mathbf{x}_i) \rvert, \label{eq:loss}
\end{align}
following from the change-of-variables theorem~\citep{tabak2013family}.
The symbol $p_z$ denotes the density function of an arbitrary $k$-dimensional distribution (e.g., an isotropic Gaussian), $f_\omega:\mathbb{R}^k \times \mathcal{X}\rightarrow \mathbb{R}^k$ denotes a continuous function invertible for its first argument $\phi$, parameterized by a neural network, and $\lvert J_{f_\omega} \rvert$ denotes the absolute value of the Jacobian's determinant of $f_\omega$ with respect to its first argument. In addition to density evaluation, as in \eqref{eq:loss}, the NF enables sampling from the modeled distribution by inverting the function $f_\omega$. 

In our experiments, we combine a convolutional neural network encoding the observations $\mathbf{x}$ with a three-step autoregressive affine NF~\citep{papamakarios2017masked} which offers a good balance between expressivity and sampling efficiency as demonstrated in \cite{wehenkel2020you}. These models have an inductive bias towards simple density functions~\citep{verine2023expressivity}, which support that the multi-modality and diversity of posterior distributions observed in the population is not an artifact of our analysis but follows from the 1D cardiovascular model and prior considered. We provide additional details on the parameterization of $f$ and the sampling algorithm in \appref{app:NF}.

\paragraph{Uncertainty analysis with SBI.} \label{sec:metrics}
Uncertainty analysis~\citep{sacks1989design, hespanhol2019understanding} regards identifiability as a continuous attribute of a model which allows ranking models by how much information the modeled observation process carries about the parameter of interest. We move away from the classical notion of statistical identifiability -- convergence in probability of the maximum likelihood estimator to the actual parameter value -- because this binary notion is not always relevant in practice and mainly applies to studies in the large sample size regime. In contrast, uncertainty analysis directly relates to the mutual information between the parameter of interest and the observation as expressed by the model considered. It captures that biased or noisy estimators are informative and may suffice for downstream tasks. 

As is standard in Bayesian uncertainty analyses, we look at credible regions $\Phi_\alpha(\mathbf{x})$ at different levels $\alpha$, which are directly extracted from the posterior distribution $p(\phi \mid \mathbf{x})$. Formally, a credible region is a subset, $\Phi_\alpha$, of the parameter space $\Phi$ over which the conditional density integrates to $\alpha$, i.e., $\Phi_\alpha: \int_{\phi \in \Phi_\alpha(\mathbf{x})} p(\phi \mid \mathbf{x}) \text{d}\phi = \alpha, \Phi_\alpha \subseteq \Phi$. In this paper, we consider the smallest covering union of regions, denoted by $\Phi_\alpha$, which is always unique in our case and in most practical settings. 

\paragraph{Size of credible intervals (SCI).}
We rely on the SCI to shed light on the uncertainty of a parameter given a measurement process. 
The SCI at a level $\alpha$ is the expected size of the credible region at this level: $\mathbb{E}_{\mathbf{x}}[ \| \tilde{\Phi}_{\alpha}(\mathbf{x}) \| ]$, where $\| \cdot \|$ measures the size of a subset of the parameter space. In practice, we split the parameter space into evenly sized cells and count the number of cells belonging to the credible interval, as detailed in \appref{app:SCI}. As discussed in \appref{app:MI_identifiability}, there exists a relationship between SCI and mutual information~(MI). However, SCI is easier to interpret for domain experts than MI, as the former is expressed in the parameter's units. In addition, SCI is robust to multi-modality in contrast to point-estimator-based metrics (e.g., mean squared/absolute error) that cannot discriminate between two posterior distributions if they lead to the same point estimate.

\paragraph{Calibration.}
Given samples from the joint distribution $p(\phi, \mathbf{x})$, credible intervals are expected to contain the true value of the parameter at a frequency equal to the credibility level $\alpha$, that is, $\mathbb{E}_{p(\phi, \mathbf{x})} \left[\mathbb{1}_{\Phi_\alpha}(\phi) \right] = \alpha $, where $\mathbb{1}$ is the indicator function. In this work, we do not have access to the true posterior but a surrogate $\tilde{p}$ of it. Hence, the coverage property of credible regions, which support the interpretation of uncertainty, may be violated, even when the forward model and the prior accurately describe the data. The calibration $C(\tilde{p}(\phi \mid \mathbf{x}), \mathcal{D})$ of a surrogate posterior $\tilde{p}$ is a metric, computed on a set $\mathcal{D}:=\{(\phi_j^\star, \mathbf{x_j})\}_{j=1}^N$, that measures whether the surrogate's credible regions respect coverage. We compute calibration as
$$
C(\tilde{p}(\phi \mid \mathbf{x}), \mathcal{D}) = \frac{1}{k}\sum_{i=1}^k \bigg| \frac{i}{k} - \frac{1}{N}\sum_{j=1}^N  \mathbb{1}_{\tilde{\Phi}_{\frac{i}{k}}}(\phi^\star_j(\mathbf{x}_j)) \bigg|,
$$
where $\tilde{\Phi}_{\frac{i}{k}}(\mathbf{x}_j)$ is the credible region at level $\alpha=\frac{i}{k}$ corresponding to the surrogate posterior distribution $\tilde{p}(\phi \mid \mathbf{x}_j)$.
The calibration directly relates to how much the surrogate posterior model violates the coverage property over all possible levels $\alpha \in \left] 0, 1 \right]$. \appref{app:calibration} describes the computation of calibration for NF-based surrogate posterior distributions.

\subsection{Model misspecification in Bayesian inference}\label{sec:misspecification_Bayesian}
Bayesian methods approach misspecification by separating the model, thus its misspecification, into two distinct components: 1) the prior distribution $p(\phi)$ and 2) the likelihood $p(\mathbf{x} \mid \phi)$. 
This division enables a focused examination of the misspecification in each component separately.

\paragraph{Prior misspecification.}
The prior distribution used to generate artificial data corresponds to a healthy population. In contrast, we consider real-world records of patients at intensive care units (ICUs) from the MIMIC~\citep{johnson2016mimic} dataset. Although the gap between the parameter distribution of the two populations is non-negligible, prior misspecification do not impair all conclusions drawn from the inspection of a model. Prior misspecification becomes insignificant if the prior support contains the real-world population or the observation carries a lot of information about the parameter -- that is, if the model's likelihood function is sharp. Thus, results gathered from the analysis of the likelihood function, e.g., by comparing prior and posterior distributions such as in the uncertainty analysis of \figref{fig:identifiability_analysis}, may transfer to real-world insights even under prior misspecification.

\paragraph{Likelihood misspecification.}
A second source of misspecification, arguably the most challenging one to overcome in practice, comes from the incorrectness of the likelihood function $p(\mathbf{x} \mid \phi)$, describing the generative process mapping parameters to observations. Common sources of misspecification include numerical approximations and simplifying modeling assumptions. Although each misspecification is unique, a common strategy is to represent the misspecification as an additional source of uncertainty and to add a noise model $p(\tilde{\mathbf{x}} \mid \mathbf{x})$ to account for it. When the initial model misspecification is small, simple noise models are sufficient, e.g., an additive Gaussian noise with constant variance. When the model does not accommodate substantial aspects of the real-world, however, designing the noise model is challenging. Moreover, adding noise has consequences as insights obtained from noisy observations must, then, rely on features that are unaltered by the noise considered. As noise increases, the number of such features shrinks. Thus, there is a balance between the amount of noise, which improves the robustness to misspecification, and the ability to obtain insights relying on complex relationships between the parameters and observations as described by the original model. As we are agnostic of the most appropriate noise model, we have considered various levels of noise in our experiments.

\section{Conclusion}

% We have demonstrated that SBI enables fine-grained uncertainty analysis and comparison of the relationship between various biomarkers and biosignals. 
We have introduced a simulation-based inference methodology to analyze complex cardiovascular simulators. 
% Our results highlight that the simulation-based inference methodology is able to extract insights consistent with variance-based sensitivity analysis. 
Our results show that our simulation-based inference method yields additional insights about 1D hemodynamics models, beyond the commonly-used VBSA and MLBSA techniques. This is done by considering the complete posterior distribution, which provides a consistent and multi-dimensional quantification of uncertainty for individual measurements.
% Furthermore, we show how simulation-based inference goes beyond variance-based sensitivity analysis owing to its fine-grained representation of uncertainty. 
This uncertainty representation enables us to recognize ambiguous inverse solutions, study the heterogeneity of sensitivity in the population considered, and understand dependencies between biomarkers in the inverse problem. Supported by results on real-world data, we have discussed the challenge of model misspecification in scientific inquiry and how to tackle it in the context of full-body hemodynamics. 
In summary, simulation-based inference enables scientists to address inverse problems in CV models, accounting for complex forward model dynamics and individualized uncertainty. Our work provides foundations for a more effective use of CV simulations for scientific inquiry and personalized medicine.

% In summary, simulation-based inference allows scientists to study inverse problems arising from CV models with a methodology that acknowledges the forward model complexity and enables an individualised representation of uncertainty. These two aspects are necessary to fully leverage CV simulations for scientific enquiry and personalised medicine.

% facilitates iterations over the scientific loop and, thus, the discovery of real-world insights from CV simulations.
% The rigour and versatility of the SBI framework should eventually enhance the scientific and real-world impact of complex cardiovascular simulations.

\bibliography{main}
\bibliographystyle{icml2023}
% \bibliography{biblio}

\clearpage

\appendix
\onecolumn

\section{Supplementary materials}
\subsection{In-vivo vs in-silico data}\label{app:sample_generation}
\begin{figure}[h]
    \centering
    \includegraphics[width=1.\textwidth]{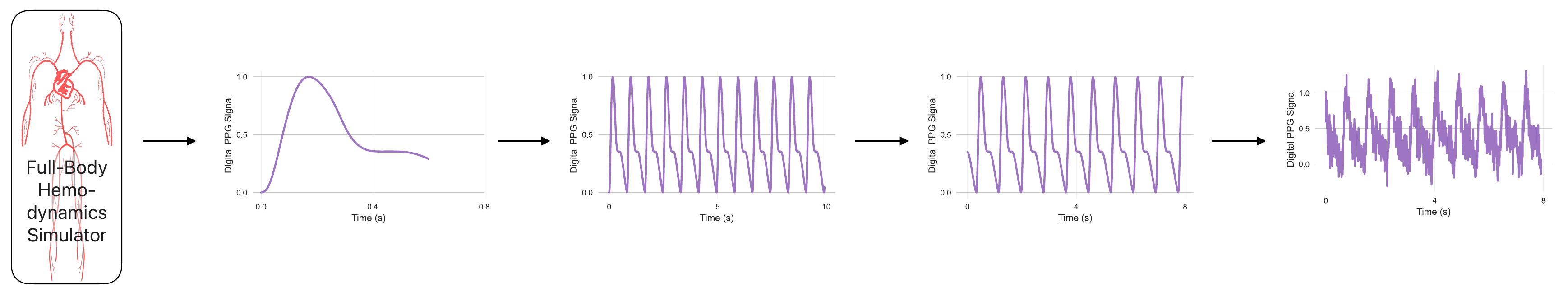}
    \caption{Generation of a digital PPG observation in-silico. From left to right: a PPG signal is extracted from the 1D hemodynamics simulator, the same wave is concatenated to reach a length of 10 seconds, the 10-second segment is cropped randomly by two seconds, additive Gaussian noise is added (SNR $\approx 11$dB).}
    \label{fig:data_gen}
\end{figure}

\begin{figure}[h]
    \centering
    \includegraphics[width=.75\textwidth]{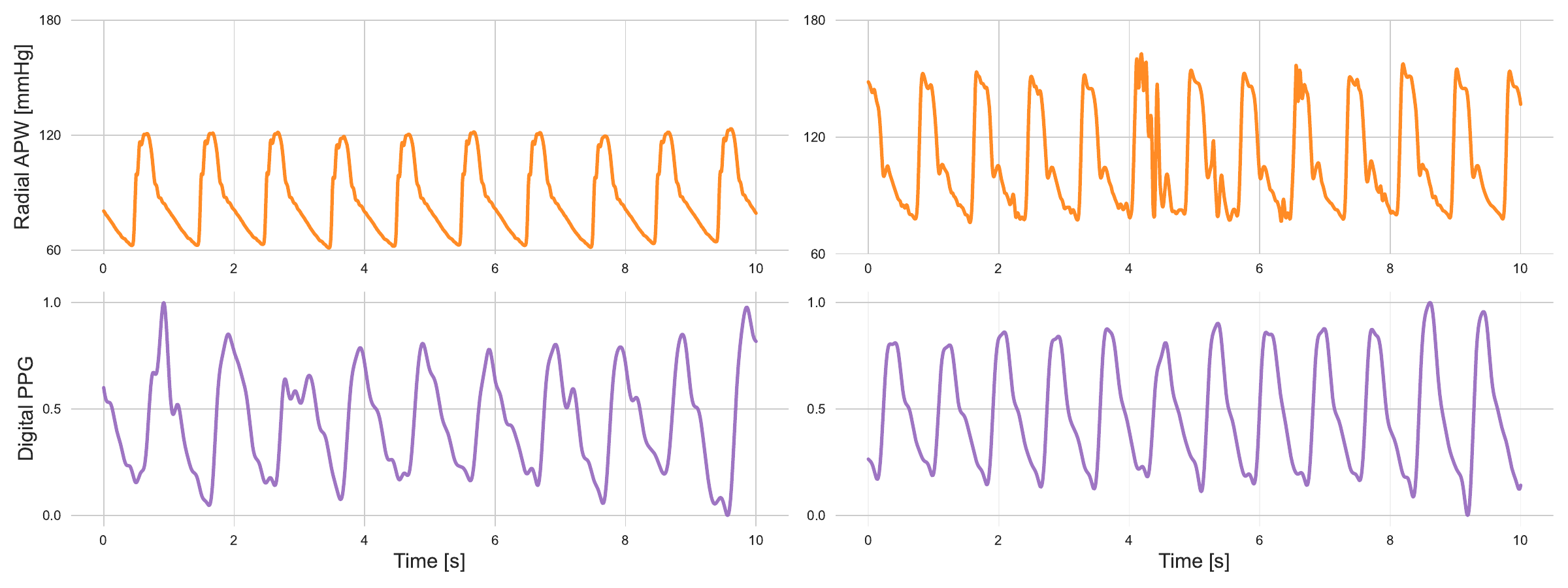}
    \caption{Waveforms reproduced from the MIMIC-III waveform database~\citep{moody2020mimic} subset of the PulseDB dataset~\citep{wang2023pulsedb}.}
    \label{fig:MIMIC}
\end{figure}

In this section we provide an overview of the generation of in-silico data, in \figref{fig:data_gen}, and a few examples of the real-world data considered from MIMIC, in \figref{fig:MIMIC}. We observe that real-world data contains degenerated beats. Moreover, another source of variation between beats comes from physiological parameter dynamics, which can vary from one beat to another. These observations motivate the introduction of noise on top of the deterministic simulation as discussed in the main paper and shown in \figref{fig:data_gen}.

\subsection{Supplementary results} 
In this section we provide additional perspectives on the learned surrogate models of the posterior distributions.
\subsubsection{Calibration and MAE}\label{app:calib_mae}
\begin{figure}[h]
    \centering  \includegraphics[width=1.\textwidth]{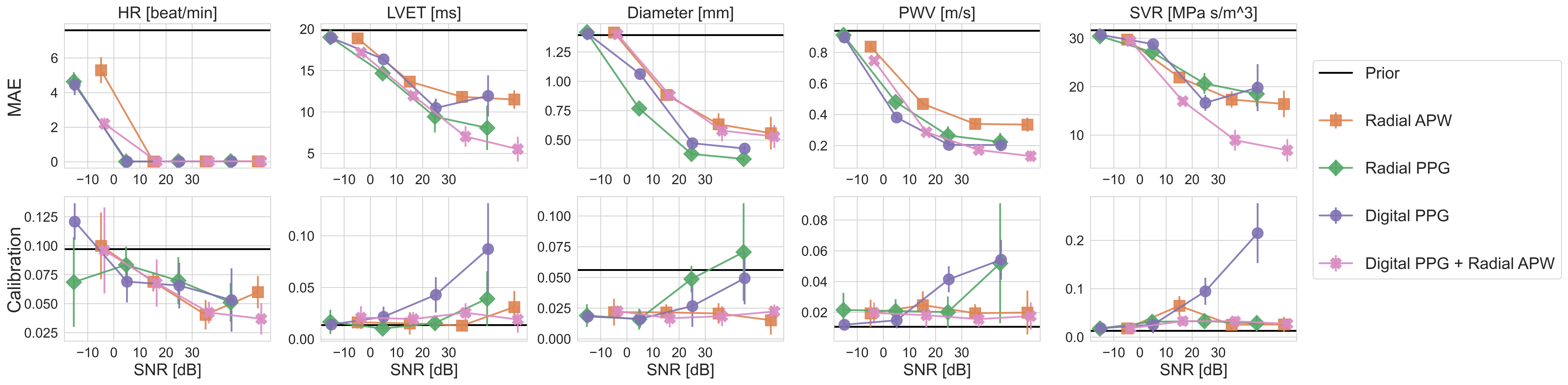}
    \caption{Mean absolute error~(MAE) of the expected value of the posterior distributions and calibration of the credible intervals. Except for the Digital PPG, at high level of noise, models are well calibrated. The MAEs follow the trend expected from the analysis of the size of credible intervals in the main materials.}
    \label{fig:MAE-calibration}
\end{figure}

\figref{fig:MAE-calibration} presents the mean average precision of a point estimator obtained as the mean of the posterior distribution and the calibration of these posterior distributions. Most surrogate models trained with NPE are well calibrated. However, there remains a risk that a surrogate model is not well calibrated, such as observed for the Digital PPG for inferring some of the parameters for low levels of noise. The MAEs have a similar behavior as the average sizes of credible intervals at size $68\%$ and $95\%$ discussed in the main materials.

\subsubsection{Posterior distributions}
\label{app:add_exp}
\begin{figure}
\centering
\begin{subfigure}{.5\textwidth}
  \centering
  \includegraphics[width=.9\linewidth]{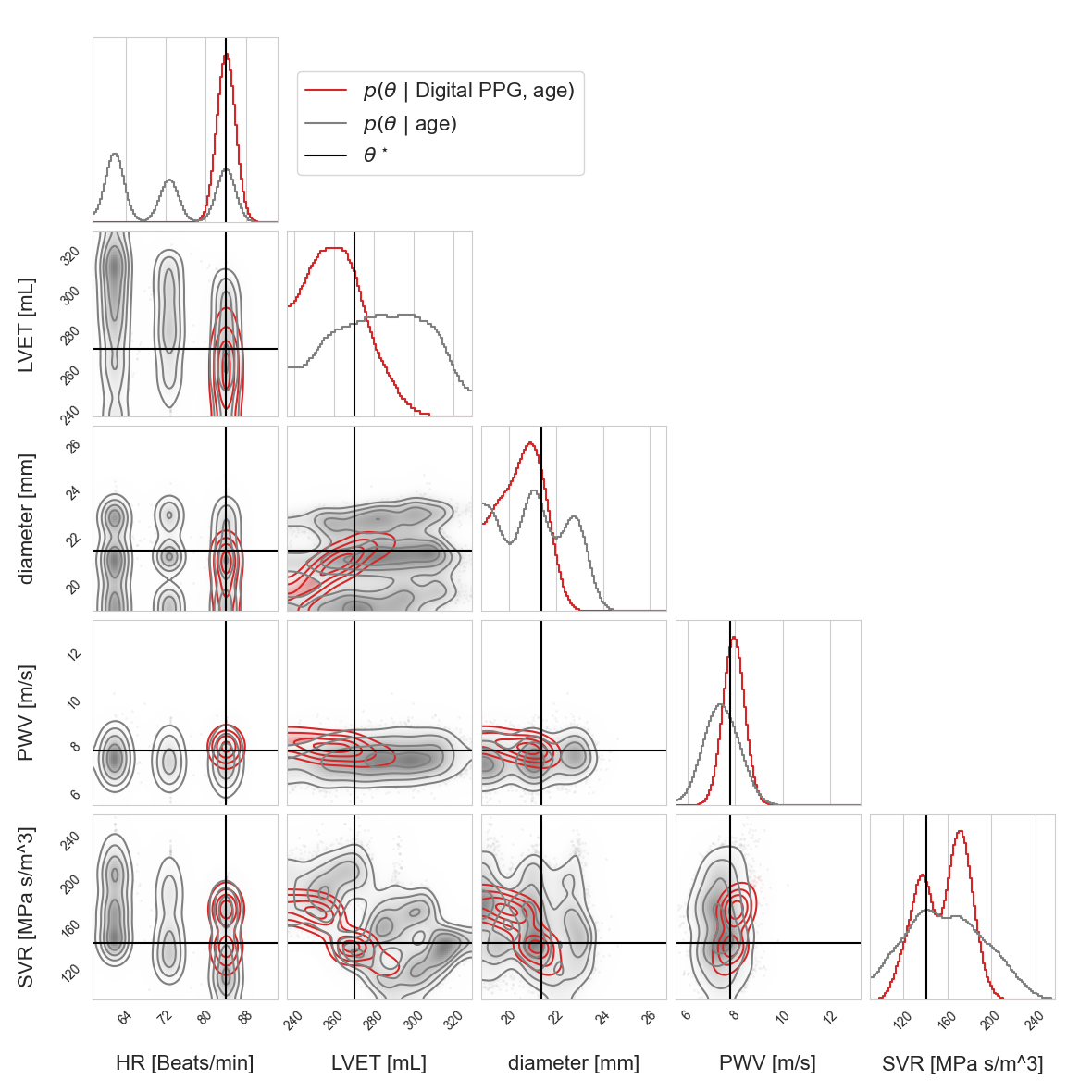}
  \caption{}
  \label{fig:sub1}
\end{subfigure}%
\begin{subfigure}{.5\textwidth}
  \centering
  \includegraphics[width=.9\linewidth]{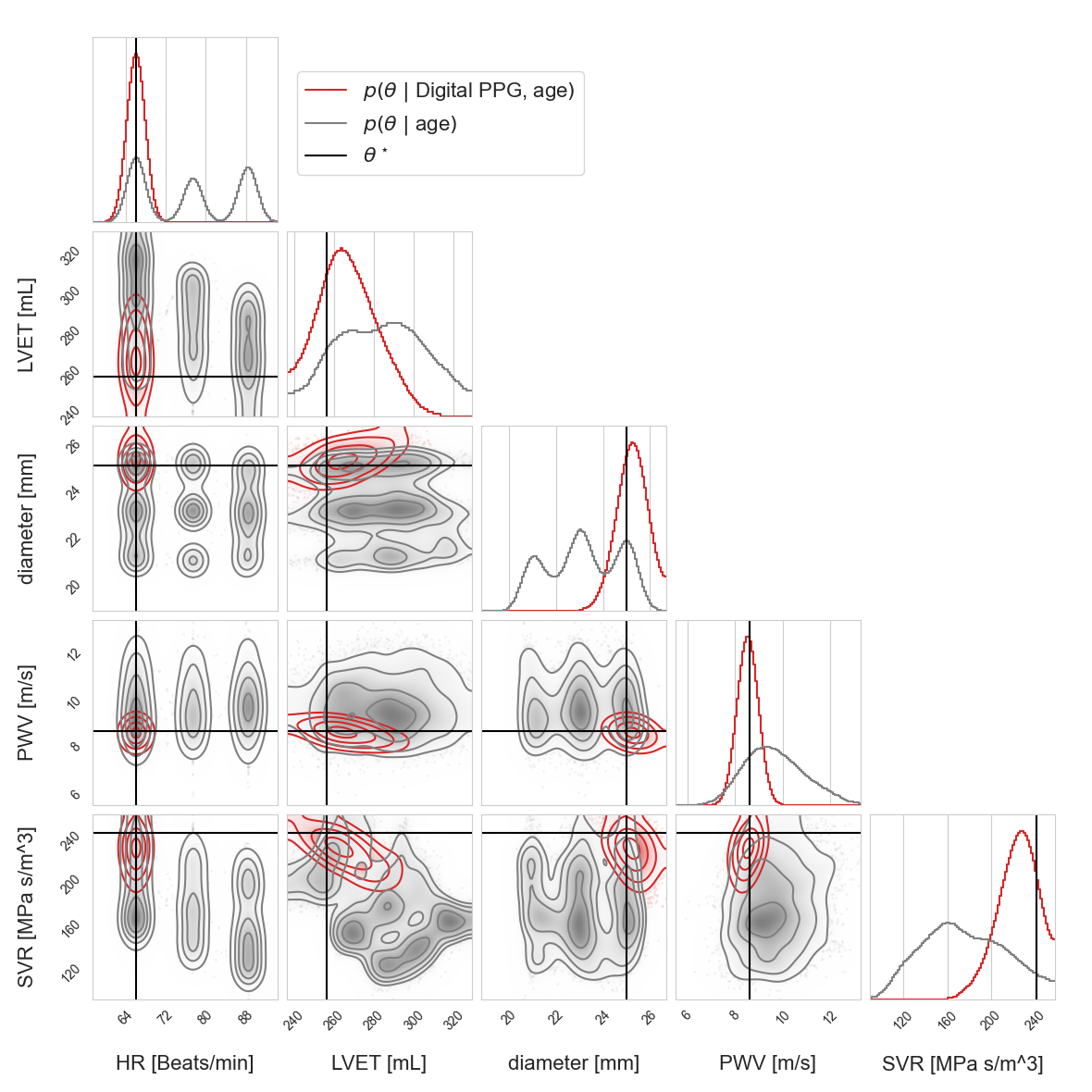}
  \caption{}
  \label{fig:sub2}
\end{subfigure}
\caption{Posterior distributions obtained for two different test observation of digital PPGs.}
\label{fig:digital_PPG_posterior}
\end{figure}

\begin{figure}
\centering
\begin{subfigure}{.5\textwidth}
  \centering
  \includegraphics[width=.9\linewidth]{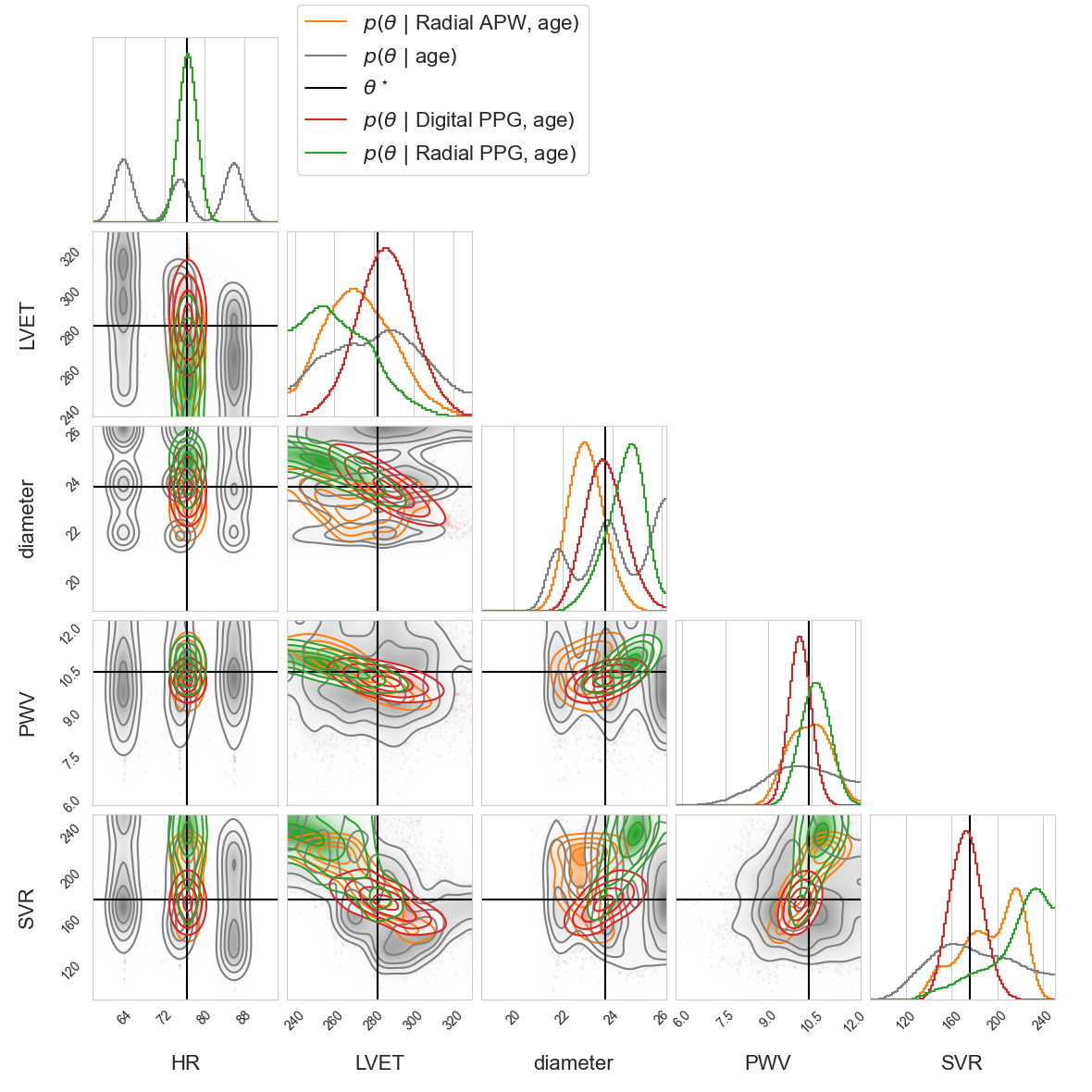}
  \caption{}
  \label{fig:sub1}
\end{subfigure}%
\begin{subfigure}{.5\textwidth}
  \centering
  \includegraphics[width=.9\linewidth]{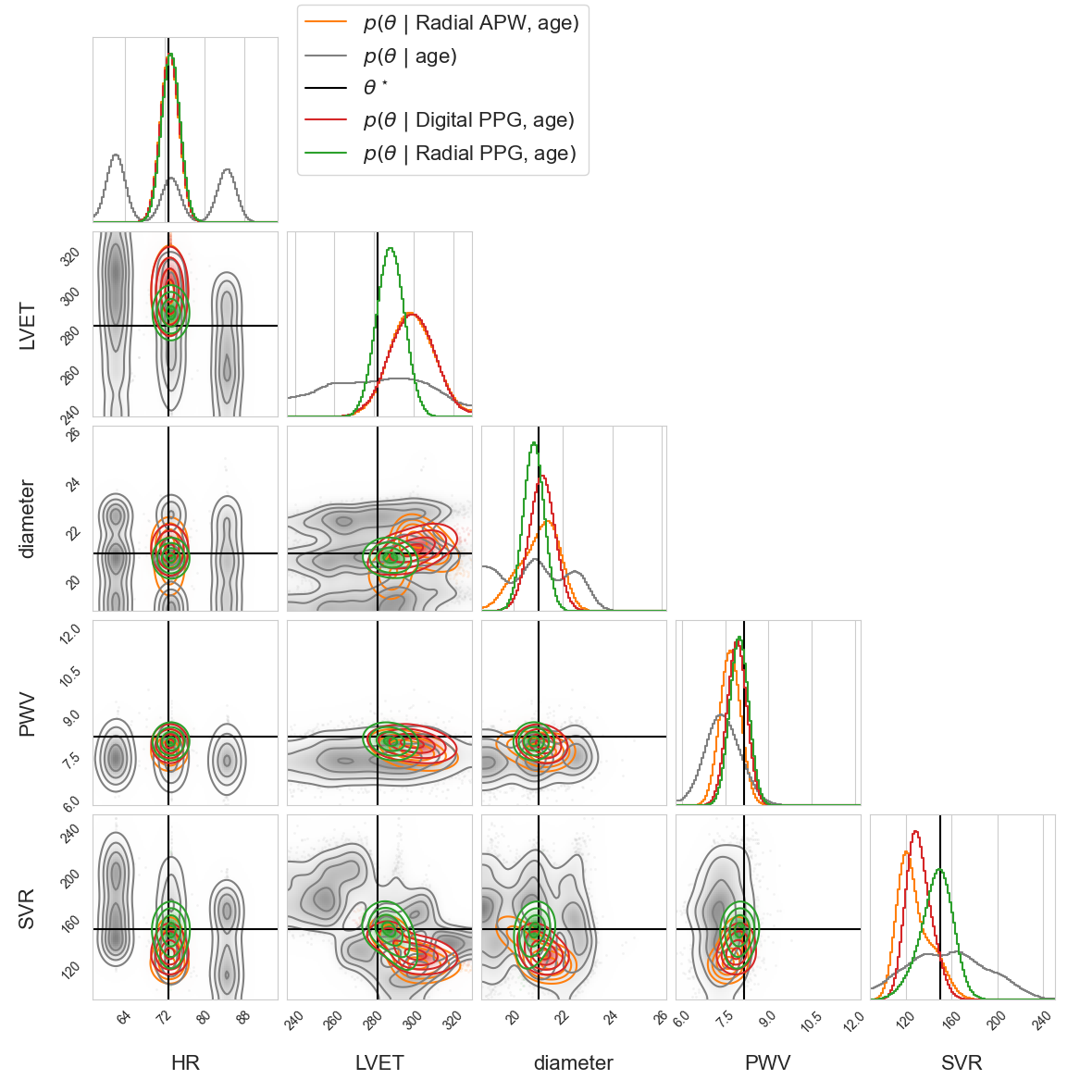}
  \caption{}
  \label{fig:sub2}
\end{subfigure}
\caption{Comparison between the posterior distributions corresponding to different measurements.}
\label{fig:all_posteriors}
\end{figure}

One desirable consequence of relying on SBI to analyze hemodynamics models is to provide access to the joint conditional distribution of parameters given an observation. In \figref{fig:digital_PPG_posterior}, we show the posterior distributions corresponding to two randomly selected simulated digital PPGs of the test set. These plots reveal how different measurements can lead to very different posterior distributions and highlight the relevance of considering sub-groups rather than the entire population at once.
In addition, \figref{fig:all_posteriors} presents the posterior distributions corresponding to the different measurement types studied. From this figure we observe that different measurements carry different information about the parameters and thus can lead to very different posterior distributions. Such plots may also indicate when multiple measurements should be done and when this is likely useless. 

\subsection{Metrics}
In this section, we provide the algorithms used to compute the calibration, in Algorithm~\ref{algorithm:statistical_calibration}, and size of credible intervals from samples of the surrogate posterior distributions, in Algorithm~\ref{algorithm:average_credible_interval}. 

\subsubsection{Calibration}\label{app:calibration}

Algorithm~\ref{algorithm:statistical_calibration} returns the distribution of minimum credibility levels required to not reject the true value of $\phi$. Under calibration, these values should be uniformly distributed -- we expect to reject falsely the true value with a frequency equal to the credibility level chosen. We report the integral (along credibility levels $\alpha$) between the observed cumulative distribution function~(CDF) of minimum credibility levels and the CDF of a uniform distribution. This metric equals $0$ under perfect calibration and is bounded by $0.5$. We report the calibration for each dimension independently as the metric does not generalize to multiple dimension.

\begin{algorithm}
\caption{Statistical Calibration of Posterior Distribution}
\label{algorithm:statistical_calibration}
\textbf{Input:} Dataset of pairs $\mathcal{D} = \{(\phi_i, x_i)\}$, Posterior distribution $p(\phi|x)$, Number of samples $N$.\\
\textbf{Output:} Distribution of minimum credibility levels.

\begin{algorithmic}[1]
\STATE Initialize an empty list $CredLevels$
\FOR{$(\phi_i, x_i) \in \mathcal{D}$}
    \STATE Initialize an empty list $Samples$
    \FOR{$i = 1$ to $N$}
        \STATE Sample $\phi_i$ from $p(\phi|x)$
        \STATE Append $\phi_i$ to $Samples$
    \ENDFOR
    \STATE Sort $Samples$
    \STATE Compute the rank (position in ascending order) $r$ of $\phi$ in $Samples$
    \STATE Set $CredLevel = \frac{r}{N}$
    \STATE Append $CredLevel$ to $CredLevels$
\ENDFOR
% \STATE Compute the distribution of minimum credibility levels based on $CredLevels$
% \RETURN Distribution of minimum credibility levels
\end{algorithmic}
\textbf{Return:} $CredLevels$.
\end{algorithm}

\subsubsection{Size of credible intervals}\label{app:SCI}
Algorithm~\ref{algorithm:average_credible_interval} describes a procedure to compute the size of credible intervals. In our experiments, we consider each dimension independently and discretize the space of value in $100$ cells. We finally report the average number of cells multiplied by the size of one cell in the physical unit of the parameter.
\begin{algorithm}
\caption{Compute Average Size of Credible Intervals}
\label{algorithm:average_credible_interval}
\textbf{Input:} Dataset of observations $x$, Posterior distribution $p(\phi|x)$, Credibility level $\alpha$
\textbf{Output:} Average size of credible intervals

\begin{algorithmic}[1]
\STATE Initialize an empty list $CredIntSizes$
\FOR{each observation $x$ in the dataset}
    \STATE Generate samples from the posterior distribution: $\phi_{\text{samples}} =$ SampleFromPosterior($p(\phi | x)$)
    \STATE Discretize the parameter space into cells
    \STATE Initialize an empty list $CellCounts$
    \FOR{each sample $\phi$ in $\phi_{\text{samples}}$}
        \STATE Increase by one the count the cell covering $\phi$
    \ENDFOR
    \STATE Sort the $CellCounts$ list in descending order
    \STATE Append the minimum number of cells required to reach the credible level $\alpha$ to $CredIntSizes$.
\ENDFOR
\STATE Compute the average size of credible intervals by taking the mean of the $CredIntSizes$ list

\end{algorithmic}
\textbf{Return:} Average size of credible intervals
\end{algorithm}
\subsubsection{Mutual information and SCI}\label{app:MI_identifiability}
\begin{figure}
    \centering
    \includegraphics[width=.8\textwidth]{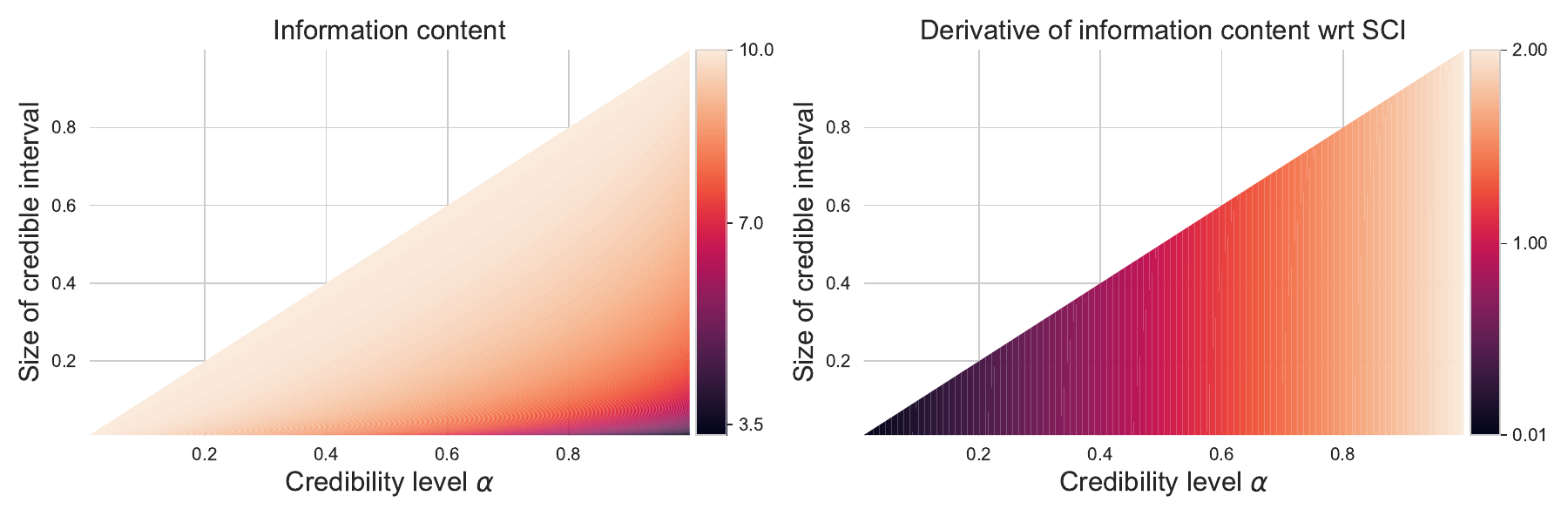}
    \caption{Relationship between the size of credible intervals and the information content present in the signal, for all credibility level. On the left: the plot of \eqref{eq:bits_SCI}; on the right: the plot of the derivative of \eqref{eq:bits_SCI} with respect to the SCI. We observe that larger SCI corresponds to larger number of bits required to encode the true value of the parameter of interest given the observation.}
    \label{fig:bits_SCI}
\end{figure}
In our experiments, we gave the average size of credible (SCI) intervals rather than the mutual information, as the former quantity is expressed in physical units that have a direct interpretation to specialists. We now discuss how the SCI relates to mutual information.

We aim to drive this discussion in the context of comparing the quality of two distinct measurement processes for inferring one quantity of interest. Formally, we denote these two measurements by $\mathbf{x}_1 \in \mathcal{X}_1$ and $\mathbf{x}_2 \in \mathcal{X}_2$ and the quantity of interest by $\phi \in \Phi$. 

Assuming a fixed marginal distribution $p(\phi)$ over the parameter, the two measurement processes $p(\mathbf{x}_1\mid \phi)$ and $p(\mathbf{x}_2 \mid \phi) $ define two joint distributions $p(
\mathbf{x}_1, \phi)$ and $p(\mathbf{x}_2, \phi) $. Considering a discretized space of parameters $\phi$, the mutual information can be written as 
\begin{align}
    \mathcal{I}(\phi, \mathbf{x}_i) = \mathcal{H}(\phi) - \mathcal{H}(\phi \mid \mathbf{x}_i),
\end{align}
 where $\mathcal{I}(\phi, \mathbf{x}_i)$ is the mutual information and $\mathcal{H}$ the entropy. As the marginal entropy of the parameter remains constant, it is clear that only the second term matters for comparing the information content of the two measurement processes. The quantity $\mathcal{H}(\phi \mid \mathbf{x}_i)$ can be interpreted as the average remaining number of bits necessary to encode $\phi$ if we know $\mathbf{x}_i$. The average is taken over the joint distribution induced by the marginal distribution of $\phi$ and the measurement process $p(\mathbf{x}_i\mid \phi)$.

From this interpretation, choosing the one with the highest mutual information is a well-motivated criterion for choosing between two measurement processes. Said differently, we are looking for measurement processes with the smallest $\mathcal{H}(\phi \mid \mathbf{x}_i)$, the one leading to small uncertainty about the correct value of $\phi$.

We use an information theory point of view to explain why, similarly to mutual information maximization~\citep{gao2020reducing}, aiming for the measurement process with the smallest SCI is a sound approach. Let us consider the measurement process $p(\mathbf{x}_i \mid 
\phi)$ leading credible intervals with size $S(\alpha, \mathbf{x}_i)$ for a certain credibility level $\alpha$ and observation $\mathbf{x}_i$. Similarly to what we do in practice to compute the SCI, we discretize the space of parameters into $N$ cells. The SCI is then defined as the minimal number of cells required to cover the credible region at level $\alpha$. From the SCI, we can say that the true value of $\phi$ belongs to one of the $S(\alpha, \mathbf{x}_i)$ cells of the credible region with probability $\alpha$ or to one of the $N - S(\alpha, \mathbf{x}_i)$ remaining cells with a probability $1-\alpha$. From this, we can bound the average number of bits required to encode the true value of $\phi$ given the observation $\mathbf{x}_i$ as a function of the SCI $S$ and the credibility level $\alpha$ as
\begin{align}
    \text{N bits} \leq -\alpha  \log_2\frac{\alpha}{S(\alpha, \mathbf{x}_i)} - (1-\alpha) \log_2 \frac{1-\alpha}{N - S(\alpha, \mathbf{x}_i)}. \label{eq:bits_SCI}
\end{align}

\figref{fig:bits_SCI} shows the relationship between this bound and the credibility level $\alpha$ and SCI. We treat SCI and the credibility $\alpha$ as independent quantities, as different measurement processes can lead to different relationships between these two quantities. We must notice that given a credibility level $\alpha$, smaller $SCI$ corresponds to better bounds. We can conclude that selecting models with the smallest SCI for a given credibility level is a sound approach with a similar interpretation as making this choice based on mutual information. 

\subsection{Whole-body hemodynamics model}~\label{app:hemodynamics_model}
\subsubsection{Parameterization}
We use a dataset of $4374$ simulations from healthy individuals aged $25$ to $75$~\citep{charlton2019modeling}. By deriving APW and simulating PPG waveforms from the blood flow and pressure data, we obtain signals corresponding to a single heartbeat of varying lengths. The parameters of interest $\phi$ can be parameters of the forward model, such as HR, LVET, Diameter, or quantity that are derived from the simulation, such as PWV and SVR. In general, the parameters of interest and the observation do not constitute a one-to-one mapping. This is especially caused by the presence of additional parameters treated as nuisance $\psi$. In the dataset from \cite{charlton2019modeling}, the following parameters vary from one simulation to the other:
\begin{itemize}
    \item Heart function:
    \begin{itemize}
        \item Heart Rate~(HR).
        \item Stroke Volume~(SV).
        \item Left Ventricular Ejection Time~(LVET). \textit{Note:} LVET changes as a deterministic function of HR and SV.
        \item Peak Flow Time~(PFT).
        \item Reflected Fraction Volume~(RFV).
    \end{itemize}
    \item Arterial properties:
    \begin{itemize}
        \item $Eh = R_{d} (k_1 e^{k_2 R_d} + k_3)$ where $k1, k2$ are constant and $k3$ follows a deterministic function of age.
        \item Length of proximal aorta
        \item Diameter of larger arteries
    \end{itemize}
    \item Vascular beds
    \begin{itemize}
        \item Resistance adjusted to achieve mean average pressure~(MAP) distribution compatible with real-world studies.
        \item Compliance adjusted to achieve realistic peripheral vascular compliance~(PVC) compatible with real-world studies.
    \end{itemize}
\end{itemize}
The interested reader will find further details in \cite{charlton2019modeling}.
\subsubsection{Measurement model}
The dataset from \cite{charlton2019modeling} is made of individual beats, which differs from real-world data usually made of a fixed-size segments. While pre-processing the real-world data to extract unique beat is feasible, it may pose challenges to ensure this extraction is consistent with the simulated waves. Instead, we add a measurement model that reduce the gap between the real-world and simulated data formats.
 We first generate segments longer than 10 seconds by concatenating the same beat multiple times. Then, we randomly crop time series into 8-second segments. This ensures that the posterior distributions are defined for 8-second segments and accounts for all possible starting positions within the heartbeat. Finally, we introduce a white Gaussian noise to the waveforms to make our analysis less sensitive to the model misspecification. \appref{app:sample_generation} showcases these steps and the resulting waveforms.

\subsection{Normalizing flows} \label{app:NF}
We provide additional details on the normalizing flows used to model the posterior distributions. In all our experiments, we apply the same training and model selection procedures. Moreover we use the same neural network architecture for all experiments.

We rely on the open-source libraries PyTorch~\citep{NEURIPS2019_9015} and \href{https://github.com/AWehenkel/Normalizing-Flows}{Normalizing Flows}, a lightweight library to build NFs built upon the abstraction of NFs as Bayesian networks from \citep{wehenkel2021graphical}. 

\subsubsection{Training setup}
We randomly divide the complete dataset into $70\%$ train, $10\%$ validation, and $20\%$ test sets. We optimize the parameters of the neural networks with stochastic gradient descent on \eqref{eq:loss} with the Adam optimizer~\citep{kingma2014adam}. We use a batch size equal to $100$, a fixed learning rate ($= 10^{-3}$), and a small weight decay ($= 10^{-6}$). We train each model for $500$ epochs and evaluate the validation loss after each epoch. The best model based on the lowest validation loss was returned and used to obtain the results presented in the paper. All data are normalized based on their standard deviation and mean on the training set. For the time series, we compute one value across the time dimension. 

\subsubsection{Neural network architecture}
We use the same neural network architecture for all the results reported. It is made of a $3$-step autoregressive normalizing flow~\citep{papamakarios2017masked} combined with a convolutional neural network~(CNN) to encode the $8$-second segments sampled at $125Hz$ ($\in \mathbb{R}^{1000}$). The CNN is made of the following layers:
\begin{enumerate}
    \item 1D Convolution with no padding, kernel size $= 3$, stride $= 2$, $40$ channels, and  followed by ReLU;
    \item 1D Convolution with no padding, kernel size $= 3$, stride $= 2$, $40$ channels, and  followed by ReLU;
    \item 1D Convolution with no padding, kernel size $= 3$, stride $= 2$, $40$ channels, and  followed by ReLU;
    \item Max pooling with a kernel $=3$;
    \item 1D Convolution with no padding, kernel size $= 3$, stride $= 2$, $20$ channels, and  followed by ReLU;
    \item 1D Convolution with no padding, kernel size $= 3$, stride $= 2$, $10$ channels, and  followed by ReLU,
\end{enumerate}
leading to a $90$ dimensional representation of the input time series. The $90$-dimensional embedding is concatenated to the $age$ and denoted $\mathbf{h}$. Then, $\mathbf{h}$ is passed to the NF as an additional input to the autoregressive conditioner~\citep{papamakarios2017masked, wehenkel2021graphical}. 

The NF is made of a first autoregressive step that inputs both the $91$ conditioning vector $\mathbf{h}$ and the parameter vector and outputs $2$ real values $\mu_i(\phi_{<i}, \mathbf{h}), \sigma_i(\phi_{<i}, \mathbf{h}) \in \mathbb{R}$ per parameter in an autoregressive fashion. Then the parameter vector is linearly transformed as $u_i = \phi_i e^{\sigma_i(\phi_{<i}, \mathbf{h})} + \mu_i(\phi_{<i}, \mathbf{h})$. The vector $\mathbf{u} := [u_1, \dots, u_k]$ is then shuffled and passed through 2 other similar transformations, leading to a vector denoted $\mathbf{z}$, which eventually follows a Gaussian distribution after learning~\citep{papamakarios2021normalizing}. The $3$ autoregressive networks have the same architecture: a simple masked multi-layer perceptron with ReLu activation functions and $3$ hidden layers with $350$ neurons each. We can easily compute the Jacobian determinant associated with such a sequence of autoregressive affine transformations on the vector $\phi$ and thus compute \eqref{eq:loss}. 

We can easily show that the Jacobian determinant is equal to the product of all scaling factors $e^{\sigma_i}$. We also directly see that ensuring these factors are strictly greater than $0$ enforce a continuously invertible Jacobian for all value of $\phi$ and thus continuous bijectivity of the associated transformation.

As mentioned, under perfect training, the mapping from $\Phi$ to $\mathcal{Z}$ defines a continuous bijective transformation that transforms samples from $\phi \sim p(\phi \mid \mathbf{h})$ into samples $\mathbf{z} \sim \mathcal{N}(0,I)$. As the transformation is bijective, we can sample from $p(\phi \mid \mathbf{h})$ by inverting the transformation onto samples from $\mathcal{N}(0,I)$. As the transformation is autoregressive, we can invert it by doing the inversion sequentially for all dimensions as detailed in \citep{papamakarios2021normalizing, wehenkel2021graphical, papamakarios2017masked}.

\end{document}